\documentclass{article} 

\usepackage{colm}          
\usepackage[utf8]{inputenc} 
\usepackage[T1]{fontenc}   
\usepackage{microtype}      
\usepackage{url}            
\usepackage{xspace}        
\usepackage[shortlabels]{enumitem} 

\usepackage[table,dvipsnames]{xcolor} 
\usepackage{color}
\usepackage{nicematrix}
\usepackage{arydshln}
\definecolor{citecolor}{HTML}{2980b9}
\definecolor{linkcolor}{HTML}{c0392b}
\definecolor{darkorange}{HTML}{FF8C00}
\definecolor{chocolate}{HTML}{D2691E}
\definecolor{darkgreen}{HTML}{006400}
\definecolor{darkblue}{HTML}{00008B}
\definecolor{mediumblue}{HTML}{0000CD}
\definecolor{dodgerblue}{HTML}{1E90FF}
\definecolor{royalblue}{HTML}{4169E1}
\definecolor{shadecolor}{RGB}{237,237,237}
\definecolor{backred}{RGB}{255, 190, 190}
\definecolor{backblue}{RGB}{210, 230, 250}
\definecolor{zrrgreen}{HTML}{008000}
\definecolor{zrrblue}{HTML}{4682B4}
\definecolor{zrrred}{HTML}{B22222}
\definecolor{rowcolor_purple}{RGB}{240, 240, 255}  
\definecolor{rowcolor_blue}{RGB}{232, 213, 233}  
\definecolor{rowcolor_green}{RGB}{240, 255, 245}  

\definecolor{highlight_purple}{RGB}{238, 232, 250} 
\definecolor{highlight_blue}{RGB}{232, 240, 255}

\definecolor{lightgray}{rgb}{.9,.9,.9}
\definecolor{darkgray}{rgb}{.4,.4,.4}
\definecolor{purple}{rgb}{0.65, 0.12, 0.82}

\usepackage{amsmath}
\usepackage{amssymb}
\usepackage{amsfonts}      
\usepackage{nicefrac}     
\usepackage{pifont}
\usepackage{fontawesome}
\usepackage{bbding}

\usepackage{graphicx}
\usepackage{wrapfig}
\usepackage{float}
\usepackage{caption}
\usepackage{adjustbox}
\usepackage{tikz}
\usepackage{tcolorbox}
\tcbuselibrary{skins}

\usepackage{longtable}

\makeatletter
  \newcommand\figcaption{\def\@captype{figure}\caption}
  \newcommand\tabcaption{\def\@captype{table}\caption}
\makeatother

\usepackage[nosfdefault]{comicneue} 
\usepackage{booktabs}     
\usepackage{multirow}
\usepackage{tabularx}
\usepackage{makecell}
\usepackage{array}

\newcolumntype{C}[1]{>{\centering\arraybackslash}m{#1}}

\usepackage{algorithm}
\usepackage{algorithmic}
\usepackage{listings}
\usepackage{subcaption}
\usepackage{CJKutf8}
\usepackage{wasysym}

\lstdefinelanguage{JavaScript}{
  keywords={break, case, catch, continue, debugger, default, delete, do, else, false, finally, for, function, if, in, instanceof, new, null, return, switch, this, throw, true, try, typeof, var, void, while, with},
  morecomment=[l]{//},
  morecomment=[s]{/*}{*/},
  morestring=[b]',
  morestring=[b]",
  ndkeywords={class, export, boolean, throw, implements, import, this},
  keywordstyle=\color{blue}\bfseries,
  ndkeywordstyle=\color{darkgray}\bfseries,
  identifierstyle=\color{black},
  commentstyle=\color{purple}\ttfamily,
  stringstyle=\color{red}\ttfamily,
  sensitive=true
}

\def\dsname{DanQing}

\newcommand{\huggingface}{\raisebox{-1.5pt}{\includegraphics[height=1.05em]{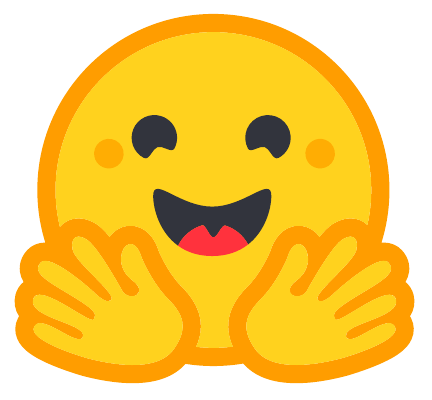}}\xspace}
\newcommand{\modelscope}{\raisebox{-1.5pt}{\includegraphics[height=1.05em]{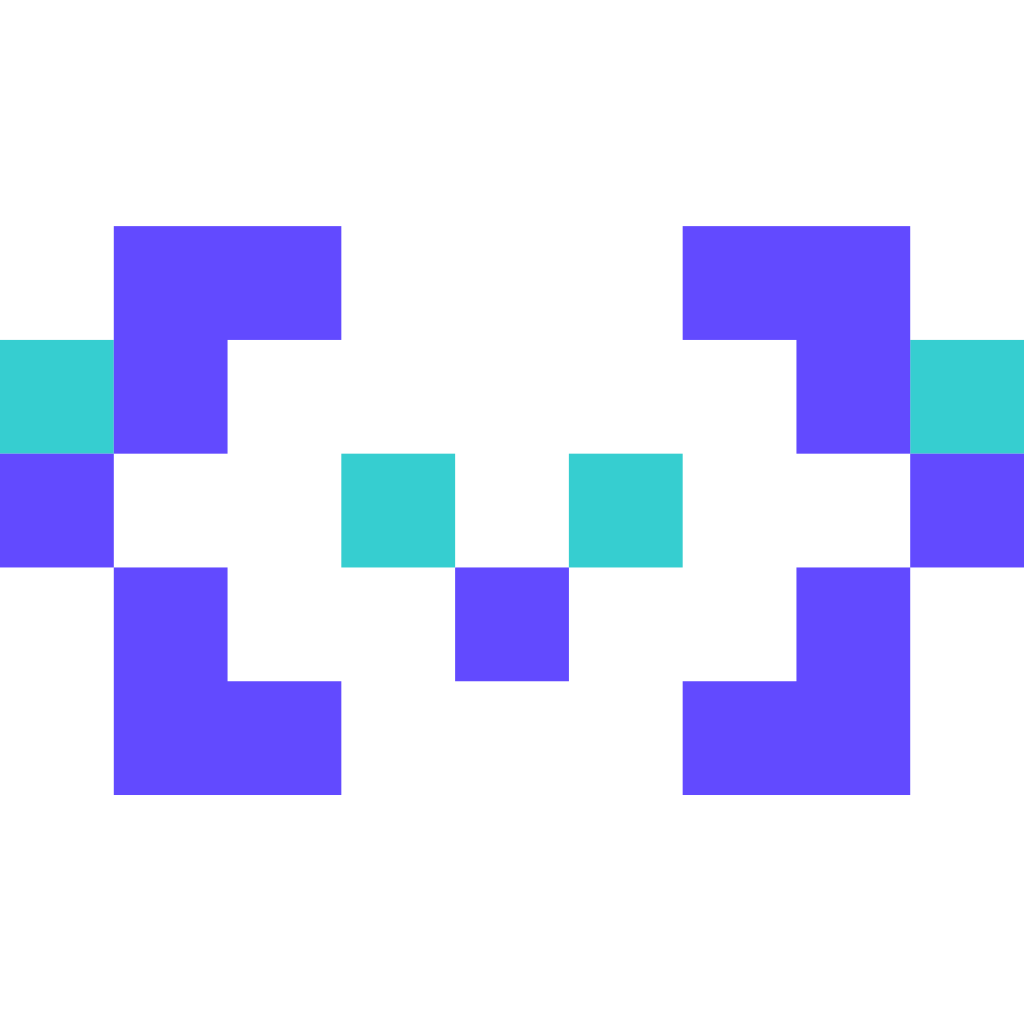}}\xspace}
\newcommand{\github}{\raisebox{-1.5pt}{\includegraphics[height=1.05em]{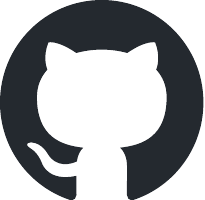}}\xspace}
\newcommand{\project}{\raisebox{-1.5pt}{\includegraphics[height=1.05em]{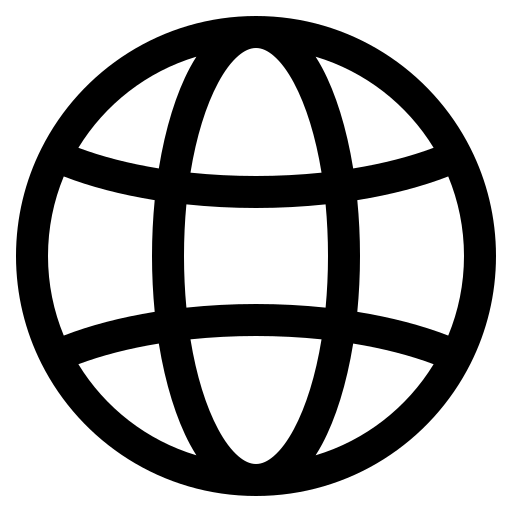}}\xspace}

\newcommand\blfootnote[1]{%
  \begingroup
  \renewcommand\thefootnote{}%
  \footnote{\noindent #1}%
  \addtocounter{footnote}{-1}%
  \endgroup
}
\makeatletter
\renewcommand{\@makefntext}[1]{\noindent\@makefnmark\ #1}
\makeatother

\usepackage[
    hidelinks,
    breaklinks=true,
    colorlinks=true,
    bookmarks=false,
    linkcolor=linkcolor,
    citecolor=citecolor,
    urlcolor=blue,
    anchorcolor=blue
]{hyperref}

\begin{document}

\title{\includegraphics[width=1.47cm, height=1cm]{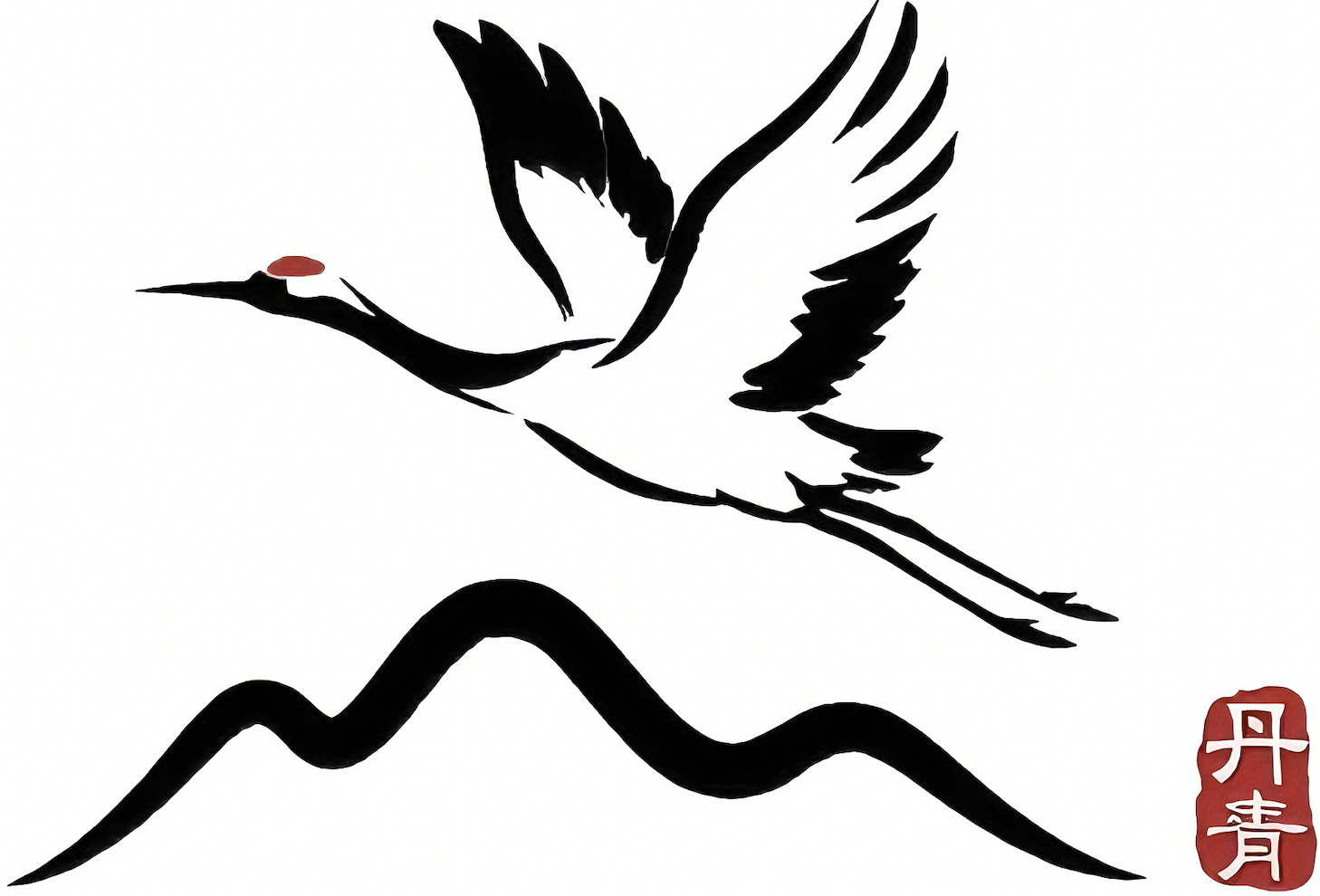}~\textit{DanQing}: An Up-to-Date Large-Scale Chinese Vision-Language Pre-training Dataset}

\author{Hengyu Shen$^{*}$, Tiancheng Gu$^{*}$, Bin Qin, Lan Wu, Yuling Wu, Shuo Tan, Zelong Sun \\ \textbf{Jun Wang, Nan Wu, Xiang An, Weidong Cai, Ziyong Feng$^{\ddagger}$, Kaicheng Yang$^{\dagger}$}\\
\\
\textit{DanQing} Team, \includegraphics[width=0.35cm, height=0.35cm]{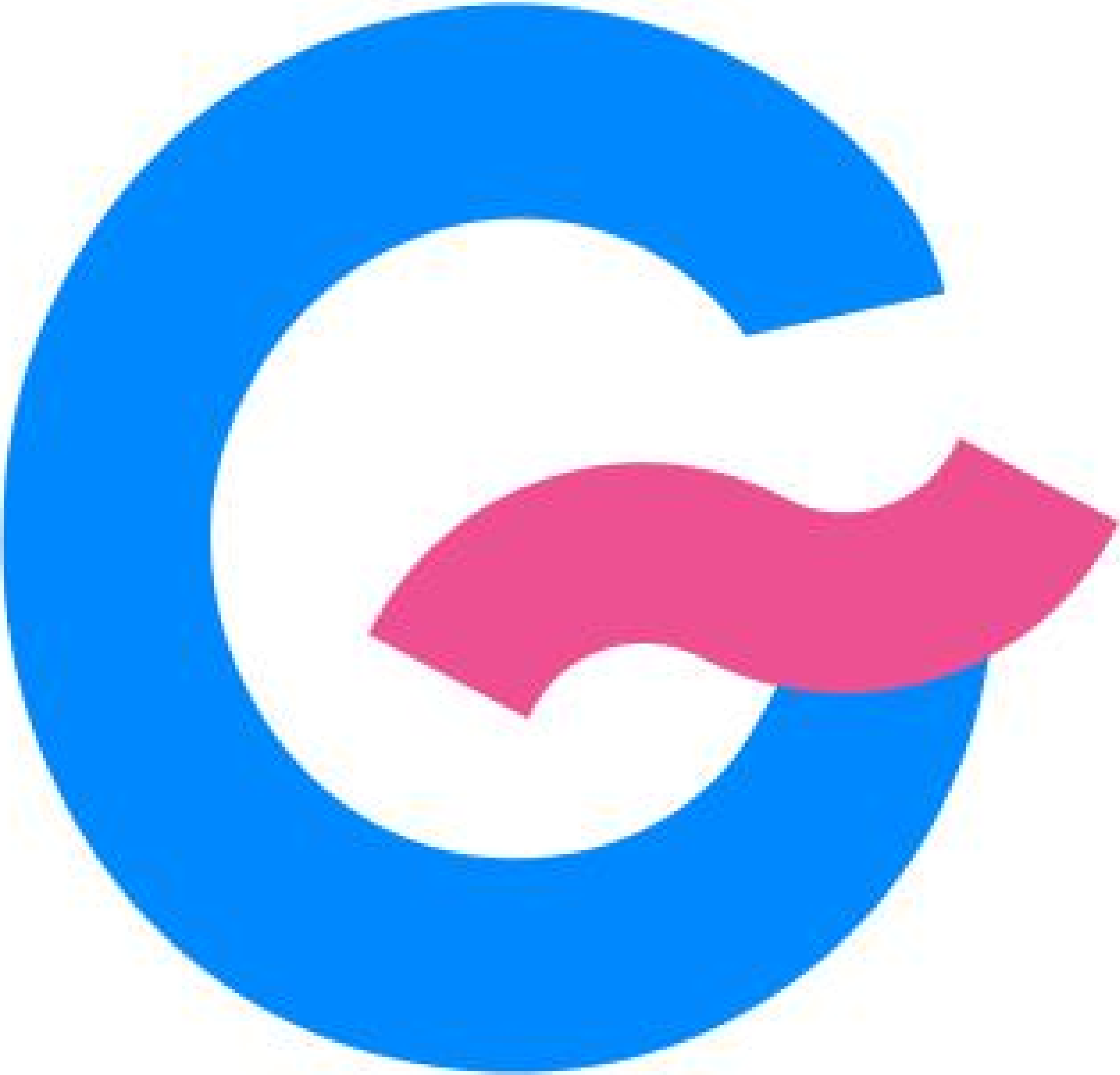} Glint Lab
}

\newcommand{\fix}{\marginpar{FIX}}
\newcommand{\new}{\marginpar{NEW}}

\colmfinalcopy

\maketitle

\begin{abstract}
Vision-Language Pre-training (VLP) models have achieved remarkable success by leveraging large-scale image-text pairs. While English-centric models like CLIP and SigLIP benefit from massive datasets (\textit{e.g.}, LAION-400M), the development of Chinese VLP remains bottlenecked by the lack of high-quality, large-scale open-source data. 
In this paper, we present \textbf{DanQing}, a large-scale Chinese cross-modal dataset containing 100 million high-quality image-text pairs curated from Common Crawl. 
To ensure superior data quality, we develop an effective systematic pipeline comprising data source selection, text refinement, visual diversification, and cross-modal cross-batch filtering, thereby effectively mitigating the intrinsic noise prevalent in web data.
Notably, DanQing incorporates data from 2024–2025, enabling models to capture contemporary semantic trends and emerging concepts. Extensive experiments via continued pretraining of SigLIP2 models demonstrate that DanQing consistently outperforms existing Chinese datasets across diverse downstream tasks, including zero-shot classification, cross-modal retrieval, and Chinese-centric large multimodal model tasks. Furthermore, in-depth analysis of DanQing reveals that it exhibits a more balanced semantic distribution and superior scaling capability compared to existing datasets.
To facilitate further research in Chinese vision-language pre-training, we will open-source the DanQing dataset under the Creative Common CC-BY-NC 4.0 license.

\end{abstract}

\vspace{-0.1cm}
\begin{center}
    \renewcommand{\arraystretch}{1}
    \resizebox{0.9\linewidth}{!}{
    \begin{tabular}{lll}
        \project & \textbf{Webpage} & {\url{https://deepglint.github.io/DanQing}} \\
        \github & \textbf{GitHub} & {\url{https://github.com/deepglint/DanQing}} \\
        \modelscope & \textbf{ModelScope} & \url{https://www.modelscope.cn/datasets/deepglint/DanQing}\\
        \huggingface & \textbf{HuggingFace} & 
        \url{https://huggingface.co/datasets/DeepGlint-AI/DanQing100M}\\
    \end{tabular}
    }
\end{center}
\vspace{-0.5cm}

\blfootnote{$^{*}$ Equal Contribution. $^{\ddagger}$ Team Leader. $^{\dagger}$ Project Leader.} 
\section{Introduction}

The proliferation of web-scale data provides a robust foundation for contrastive vision-language representation learning~\cite{gu2025realsyn}. By aligning dual-encoder architectures through image-text correspondence, frameworks like CLIP~\cite{clip} have demonstrated remarkable generalization across diverse downstream tasks including image captioning~\cite{mokady2021clipcap, li2022blip, yu2022coca}, object detection~\cite{gu2022vild, li2022glip, zhong2022regionclip}, semantic segmentation~\cite{li2022lseg, rao2022denseclip, xu2022groupvit}, and cross-modal retrieval~\cite{longclip, kempf2025and}. Given the efficacy of CLIP, this promising paradigm has garnered significant attention from both industry and academia as a potential pathway toward next-generation foundational AI models~\cite{wukong,datacomp}.

The success of Vision-Language Pre-training (VLP) is primarily driven by the synergy between architectural innovation and data scaling. On one hand, advanced modeling techniques such as ViT~\cite{ViT} and BERT~\cite{devlin2019bert}, optimized via InfoNCE~\cite{clip} or Sigmoid-based~\cite{siglip} contrastive losses, have significantly enhanced the ability of dual-tower models to learn semantically rich multimodal embeddings. On the other hand, the availability of large-scale datasets like LAION-5B~\cite{laion5b} and DataComp-1B~\cite{datacomp} has provided an essential foundation for VLP. However, while English-centric resources continue to expand, the development of Chinese image-text datasets has substantially lagged. Notably, the most recent dataset~\cite{ccmb} is introduced over three years ago. As a result, despite the recognized importance of data scale, the Chinese vision-language pretraining remains limited. Furthermore, existing Chinese resources often suffer from significant data decay, where a high proportion of inaccessible image URLs severely impairs model training and reproducibility. This scarcity of high-quality, persistent data bottlenecks Chinese cross-modal representations.

\begin{figure*}[t!]
    \centering
    \includegraphics[width=\linewidth]{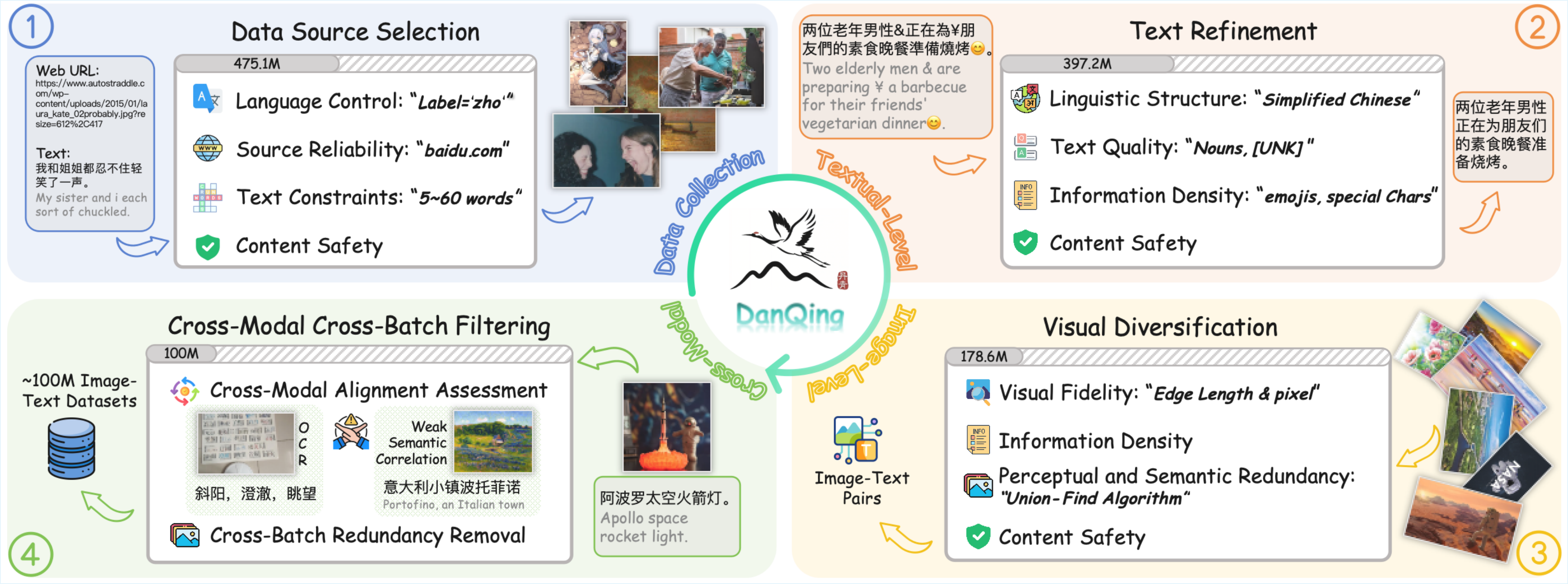}
    \vspace{-6mm}
    \caption{Overview of the DanQing dataset construction pipeline.}
    \vspace{-7mm}
    \label{fig:overview_method}
\end{figure*}

\begin{wraptable}[]{r}{0.6\textwidth}
\centering
\small
\vspace{-3mm}
\begin{minipage}[c]{0.6\textwidth}
    \centering
    \setlength{\tabcolsep}{4pt}
    \renewcommand{\arraystretch}{1.1}
    \resizebox{\linewidth}{!}{
    \begin{tikzpicture}
    \node[inner sep=2.5pt] (tbl) {
    {
    \begin{NiceTabular}{lcccccr} 
    \textbf{Dataset} & \textbf{Year} & \textbf{Language} & \textbf{Availability} & \textbf{Size} & \textbf{Success Rate} \\
    \midrule
    CC3M~\cite{cc3m} & 2018 & English & Yes & 3.1M & $\approx$60\% \\
    CC12M~\cite{cc12m} & 2021 & English & Yes & 12M & $\approx$60\% \\
    RedCaps~\cite{redcaps} & 2021 & English & Yes & 12M & - \\
    WIT~\cite{wit} & 2021 & Multilingual & Yes & 11.5M & - \\
    YFCC100M~\cite{yfcc100m} & 2014 & English & Yes & 100M & $\approx$70\% \\
    COYO~\cite{coyo700m} & 2022 & English & Yes & 700M & -\\
    LAION-400M~\cite{laion400m} & 2021 & English & Yes & 400M & - \\
    RealSyn~\cite{gu2025realsyn} & 2025 & English & Yes & 100M & - \\
    \midrule
    Product1M~\cite{product1m} & 2021 & Chinese & Yes & 1M & - \\ 
    WudaoMM~\cite{wudaomm} & 2022 & Chinese & Yes & 5M & -\\
    M6-Corpus~\cite{m6} & 2021 & Chinese & No & 60.5M & -\\
    Wukong~\cite{wukong} & 2022 & Chinese & Yes & 100M & $\approx$85\%\\
    TaiSu~\cite{taisu} & 2022 & Chinese & Yes & 166M & 100\%\\
    Zero~\cite{ccmb} & 2022 & Chinese & Yes & 250M & $\approx$60\% \\
    \rowcolor{rowcolor_blue} DanQing & 2025 & Chinese & Yes & 100M & 100\% \\
    \end{NiceTabular}
    }
    };
    
    \draw[line width=0.08pt, rounded corners=4pt]
    (tbl.south west) rectangle (tbl.north east);
    
    \draw[line width=0.08pt, rounded corners=4pt, opacity=0.6]
    ([xshift=0.5pt,yshift=0.5pt]tbl.south west)
    rectangle
    ([xshift=0.5pt,yshift=0.5pt]tbl.north east);
    \end{tikzpicture}
    }
\end{minipage}
\vspace{-2mm}
\caption{Overview of existing VLP datasets.}
\vspace{-5mm}
\label{tab:datasets_source_info}
\end{wraptable}
To bridge this gap, we present \dsname, a large-scale Chinese dataset comprising 100 million high-quality image-text pairs collected after 2024. Specifically, we develop an effective systematic pipeline (Fig.~\ref{fig:overview_method}) to refine 1 billion raw pairs into a high-quality subset. This pipeline consists of data source selection, text refinement, visual diversification, and cross-modal cross-batch filtering, each designed to address distinct sources of noise in web data. As a result, we filter out 90.46\% of the raw data, substantially reducing intrinsic noise and enhancing overall dataset quality. Extensive evaluations via continued pre-training of SigLIP2 models demonstrate that DanQing consistently outperforms existing Chinese datasets across diverse downstream tasks, including zero-shot classification, cross-modal retrieval, and Chinese-centric large multimodal model tasks. Furthermore, in-depth analysis of DanQing reveals that DanQing exhibits a more balanced semantic distribution and superior scaling capability compared to existing datasets. The main contributions of this paper are summarized as follows:
\begin{itemize}[leftmargin=*,noitemsep,topsep=2pt]
    \item We \textbf{develop an effective data filtering pipeline} tailored for processing large-scale Chinese image-text pairs obtained from the Internet.
    \item We \textbf{release an up-to-date large-scale Chinese image-text dataset \dsname}, which comprises nearly 100 million Chinese image-text pairs collected after 2024.
    \item We \textbf{conduct extensive experiments across multiple downstream tasks} to demonstrate the effectiveness and scalability of \dsname.
\end{itemize}

\section{Related Work}
\subsection{Vision-Language Pretraining}
As a seminal work in vision-language pre-training, CLIP~\cite{clip} has demonstrated exceptional zero-shot recognition and transfer capabilities. Building on this paradigm, recent studies have introduced various enhancements~\cite{clipcid,rwkvclip,wu2023grounded}. For instance, SLIP~\cite{slip} integrates self-supervised learning with image-text pretraining to improve representation quality, while ALIP~\cite{alip} employs a gating mechanism for dynamic sample reweighting to mitigate the impact of noisy data.  DFN~\cite{dfn} proposes novel data filtering networks to construct high-quality image-text datasets. Furthermore, MetaCLIP~\cite{metaclip} and MetaCLIP2~\cite{metaclip2} leverage metadata derived from CLIP’s semantic concepts to curate balanced data subsets. To enable larger batch sizes, SigLIP~\cite{siglip} and SigLIP2~\cite{siglip2} adopt a sigmoid-based loss, eliminating the need for global normalization. In the context of Chinese vision-language pre-training, ChineseCLIP~\cite{chineseclip} introduces a two-stage framework comprising locked-image tuning followed by contrastive tuning. Similarly, R2D2~\cite{ccmb} enhances representation learning through a preranking-ranking strategy combined with bidirectional distillation. Despite these advances, the scarcity of large-scale, high-quality Chinese image-text datasets bottlenecks vision-language pretraining, hindering model scalability and generalization.


\subsection{Large-Scale Image-Text Dataset}
The impressive performance of CLIP~\cite{clip} across downstream tasks is primarily attributed to the availability of massive, high-quality image-text data. To further advance model capabilities, numerous large-scale image-text pair datasets have been introduced in recent years~(Tab.~\ref{tab:datasets_source_info}). The YFCC100M~\cite{yfcc100m} dataset provides a comprehensive record of photo and video sharing trends on Flickr from its inception in 2004 through early 2014. LAION400M~\cite{laion400m} comprises 400 million image-text pairs sourced from Common Crawl and has become a standard benchmark for vision-language pretraining. COYO-700M~\cite{coyo700m} collects approximately 10 billion image-alt-text pairs from HTML documents in Common Crawl (from October 2020 to August 2021), employing efficient image- and text-level filtering to remove uninformative pairs at minimal cost. To support data filtering research and benchmarking, DataComp~\cite{datacomp} assembled a pool of 12.8 billion image-text pairs for competition tracks, model training, and evaluation. However, these datasets are predominantly based on English image-text pairs, while large-scale Chinese image-text datasets remain scarce. To address this gap, the Wukong~\cite{wukong} dataset, comprising 100 million Chinese image-text pairs collected from the web, has been released. Taisu~\cite{taisu} further advances this effort by introducing an automatic filtering framework, resulting in a large-scale, high-quality Chinese multimodal dataset containing approximately 166 million images and 219 million Chinese captions. Leveraging user click-through rates and diverse textual information for each image, the Zero~\cite{ccmb} dataset offers 250 million images and 750 million corresponding Chinese texts, significantly advancing resources for Chinese vision-language pretraining. 
Despite these advancements, existing Chinese image-text datasets still lag behind their English counterparts in both scale and quality.
\section{\dsname\ Dataset}

\subsection{Training Objective of \dsname}
The \dsname\ dataset aims to enhance the Chinese multimodal embedding capabilities of CLIP-style models. Given L2-normalized embeddings $\mathbf{v}, \mathbf{t} \in \mathbb{R}^d$ from the image encoder $f_v(\cdot)$ and text encoder $f_t(\cdot)$, we adopt the SigLIP~\cite{siglip2} objective, which reformulates alignment as independent binary classification tasks using a sigmoid loss:
\[
\mathcal{L} = - \sum_{i,j} \left[ \mathbb{I}_{i=j} \log \sigma(s_{ij}) + \mathbb{I}_{i \neq j} \log (1 - \sigma(s_{ij})) \right],
\]
where $s_{ij} = (\mathbf{v}_i \cdot \mathbf{t}_j) / \tau + b$ is the scaled similarity with bias $b$. Compared to the standard CLIP~\cite{clip} cross-entropy over a batch of $B$ image-text pairs, this logistic formulation avoids batch coupling and scales more favorably to large batches and distributed training.

\subsection{Curation of \dsname}
\begin{figure}[t!]
  \centering
  \begin{subfigure}[b]{0.49\textwidth}
    \centering
    \includegraphics[width=\textwidth]{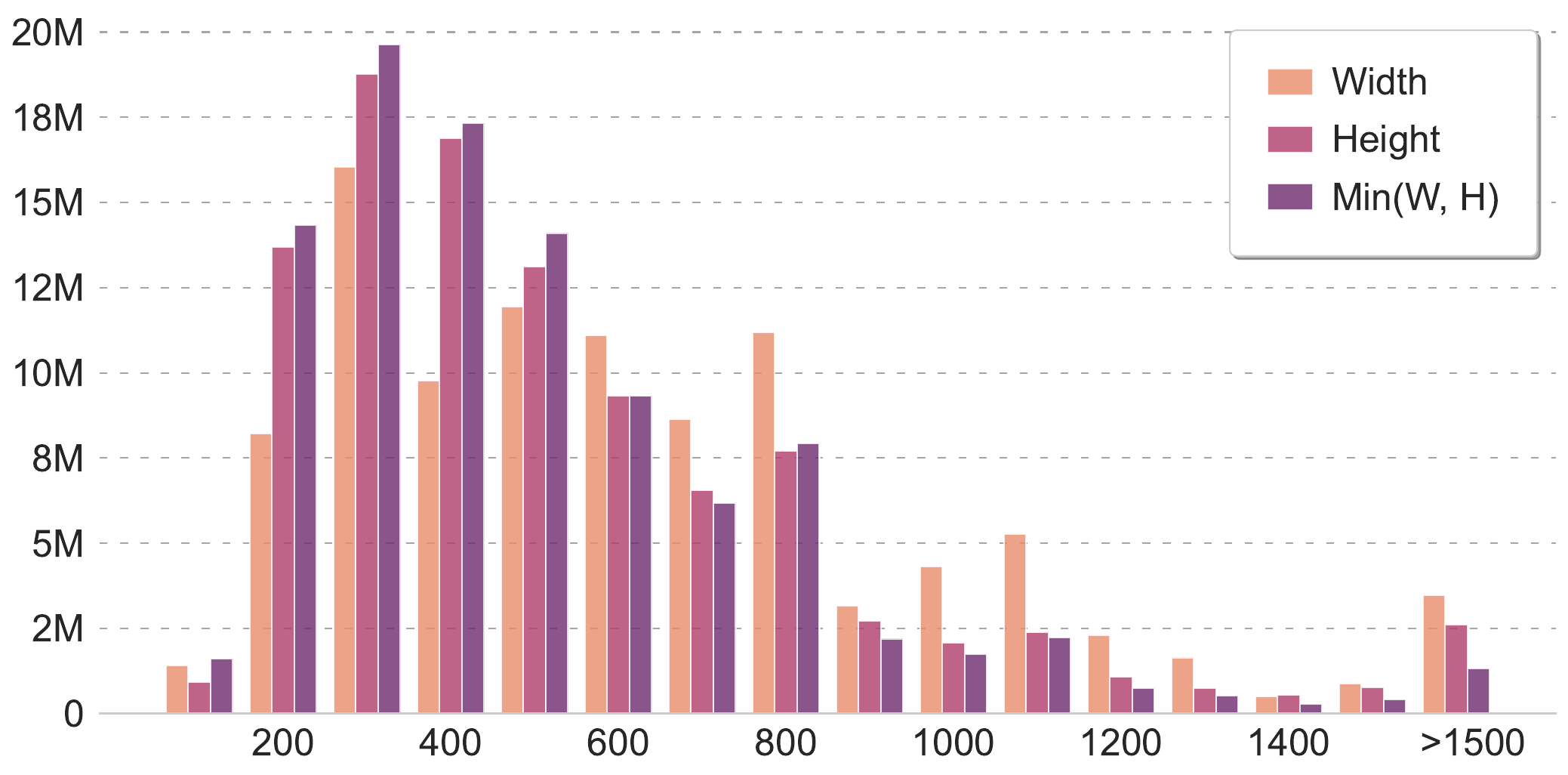}
    \vspace{-5mm}
    \caption{Image Resolution Distribution.}
    \label{fig:image_resolution}
  \end{subfigure}
  \hfill
  \begin{subfigure}[b]{0.49\textwidth}
    \centering
    \includegraphics[width=\textwidth]{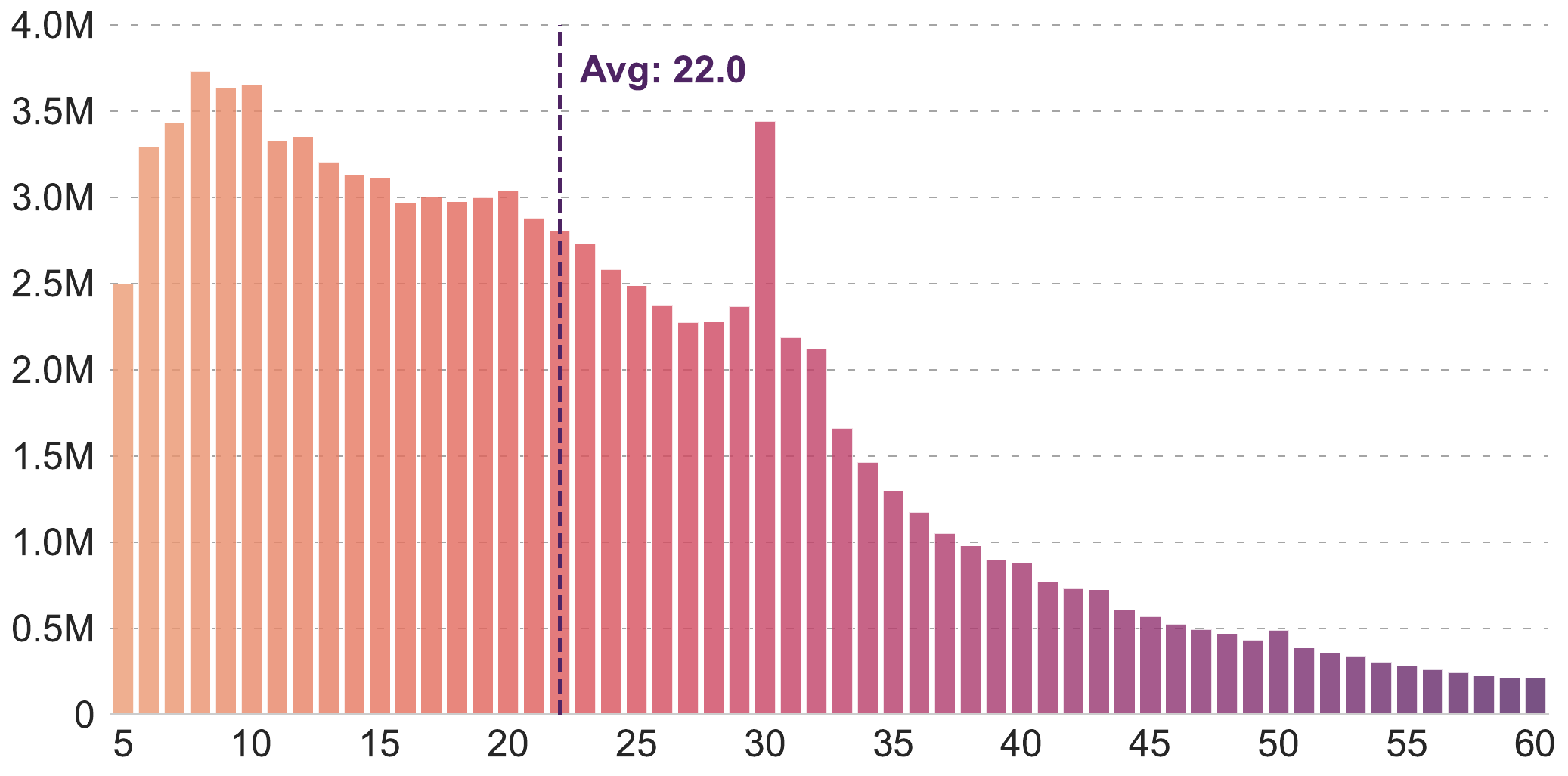}
    \vspace{-5mm}
    \caption{Text Chinese words length.}
    \label{fig:text_lenth}
  \end{subfigure}
  \hfill
  \vspace{-2mm}
  \caption{Overview of data characteristics in \dsname.}
  \vspace{-2mm}
  \label{fig:data character}
\end{figure}
\noindent\textbf{Data Source Selection.} We first collect raw image-text pairs from the Common Crawl (2024–2025). The data collection is partitioned into seven batches and processed in parallel to ensure efficiency. By filtering for the ``zho'' language tag, we obtain an initial pool of approximately 1.05 billion pairs. To mitigate the inherent noise in web-scale data, we implement a coarse-grained curation based on three criteria: {\large{\ding{172}}} \textit{Source Reliability}: We exclude the pairs originating from websites listed in a manually curated blacklist of low-quality sources. {\large{\ding{173}}} \textit{Textual Constraints}: We retain the data which text contain 5 to 60 Chinese words. {\large{\ding{174}}} \textit{Content Safety}: We use a lightweight 1M-parameter binary classifier~\cite{PaddleHubPornDetection2021} filters unsafe content. 
This stage yields 706M pairs ($\sim$67\%), and subsequent downloading achieve a 67\% success rate, resulting in 475M accessible image–text pairs.

\noindent\textbf{Text Refinement.} We implement a multi-stage text refinement pipeline across four dimensions: linguistic structure, text quality, information density, and content safety, primarily due to its lower computational overhead. \textit{{\large{\ding{172}}} Linguistic Structure:} We employ FastText~\cite{joulin2016fasttext} to identify and retain Chinese text based on language identification confidence, followed by OpenCC~\cite{OpenCC} to standardize all content to Simplified Chinese. \textit{{\large{\ding{173}}} Text Quality:} To ensure grammatical and lexical integrity, we discard samples that lack nouns or contain more than five \texttt{[UNK]} tokens after SigLIP2 tokenization~\cite{siglip2}. \textit{{\large{\ding{174}}} Information Density:} Following RealSyn~\cite{gu2025realsyn}, we remove emojis and special characters, and apply an entropy-based semantic filter $H = - \sum_{i=0}^{L} P(c_i) \log_2 P(c_i)$, where $P(c_i)$ denotes the probability of token $c_i$ and $L$ denotes the total number of words in the text. Captions with $H < 6\times10^{-4}$ are eliminated and this threshold is empirically determined to filter out low-content captions. \textit{{\large{\ding{175}}} Content Safety:} We utilize a 20M-parameter NSFW detector~\cite{PaddleHubPornDetectionCNN2021} and the Baidu DataBuilder service to filter advertisements, sensitive political content, and territorial disputes, based on their proven effectiveness in prior work. This process reduces the corpus from 475M to 397M pairs (a 16.4\% reduction), significantly improving the signal-to-noise ratio for subsequent training.

\noindent\textbf{Visual Diversification.} To rigorously ensure both perceptual quality and semantic diversity, we establish a multi-stage visual diversification pipeline that systematically addresses visual fidelity, information density, perceptual and semantic redundancy, and content safety. {\large{\ding{172}}}\textit{Visual Fidelity:} We retain images with aspect ratios between 1:3 and 3:1 and a minimum edge length greater than 100 pixels. Images with low pixel intensity variation ($\sigma < 2$), indicative of uninformative content, are removed. Blurry samples are filtered out by requiring a Laplacian variance (computed via OpenCV~\cite{opencv_library}) of at least 1000. {\large{\ding{173}}}\textit{Information Density:} Image complexity is quantified via entropy ($H = - \sum_{i=0}^{255} P(i) \log_2 P(i)$, where $P(i)$ denotes the probability of pixel value $i$) and exclude images with entropy below 3 to remove low-information samples. {\large{\ding{174}}}\textit{Perceptual and Semantic Redundancy:} To suppress duplication, inspired by previous work~\cite{clipcid}, we extract image embeddings with Chinese-CLIP-L14~\cite{chineseclip} and compute pairwise cosine distances. Images with cosine distances below an empirically determined threshold ($\beta = 0.1$) are divided into the same set using the Union-Find algorithm~\cite{tarjan1975efficiency}. Within each set, we retain only the central image (the image nearest to the centroid of the set) and only the centroid-closest image is retained per set.
{\large{\ding{175}}}\textit{Content Safety:} We apply an 86M-parameter NSFW detection model~\cite{freepik2025nsfw} to remove pairs with the highest risk scores, ensuring content safety.
Through this process, the dataset size is reduced from 397M to 178M pairs (44.8\% retention), substantially improving visual and semantic quality for downstream applications.

\noindent\textbf{Cross-Modal Cross-Batch Filtering.} Following LAION400M~\cite{laion400m}, we leverage an expert model to further refine the dataset based on image-text alignment. Specifically, we compute L2 distance using Chinese-CLIP-L14~\cite{chineseclip} and retain pairs within the [1.06, 1.24] interval. This thresholding strategy ensures high-quality alignment: scores below 1.06 indicate weak semantic correlation, while those exceeding 1.24 often correspond to images dominated by OCR text rather than descriptive content. This stage removes 25M pairs. Finally, we apply perceptual and semantic redundancy filtering to the samples across seven batches, eliminating an additional 54 million redundant pairs. As a result, this pipeline produces a curated dataset comprising 99,892,381 high-quality image-text pairs.

\subsection{Statistic of \dsname}
\label{subsec: statistic of danqing}

\noindent\textbf{Data Characteristics.} As illustrated in Fig.~\ref{fig:image_resolution}, we analyze the general characteristics of the \dsname\ dataset. We assess image resolutions in terms of width, height, and minimum dimension, revealing a broad spectrum of visual scales. While most images fall within the 300 to 500 pixel range, a considerable proportion exceeds 1,024 pixels. 
This extensive coverage facilitates the extraction of robust, scale-invariant features for vision-language representation learning. Such visual diversity is essential for ensuring generalization to real-world images, where object scales and orientations vary substantially. In addition, we present the distribution of text lengths in Fig.~\ref{fig:text_lenth}. \dsname\ contains a total of 2.2B Chinese words, with an average length of 22 words per sample. The length distribution spans a broad range from 5 to 60 tokens, with the majority concentrated between 5 and 40. This wide distribution enables models to learn representations across different levels of textual granularity.

\begin{figure*}[t!]
    \centering
    \includegraphics[width=\linewidth]{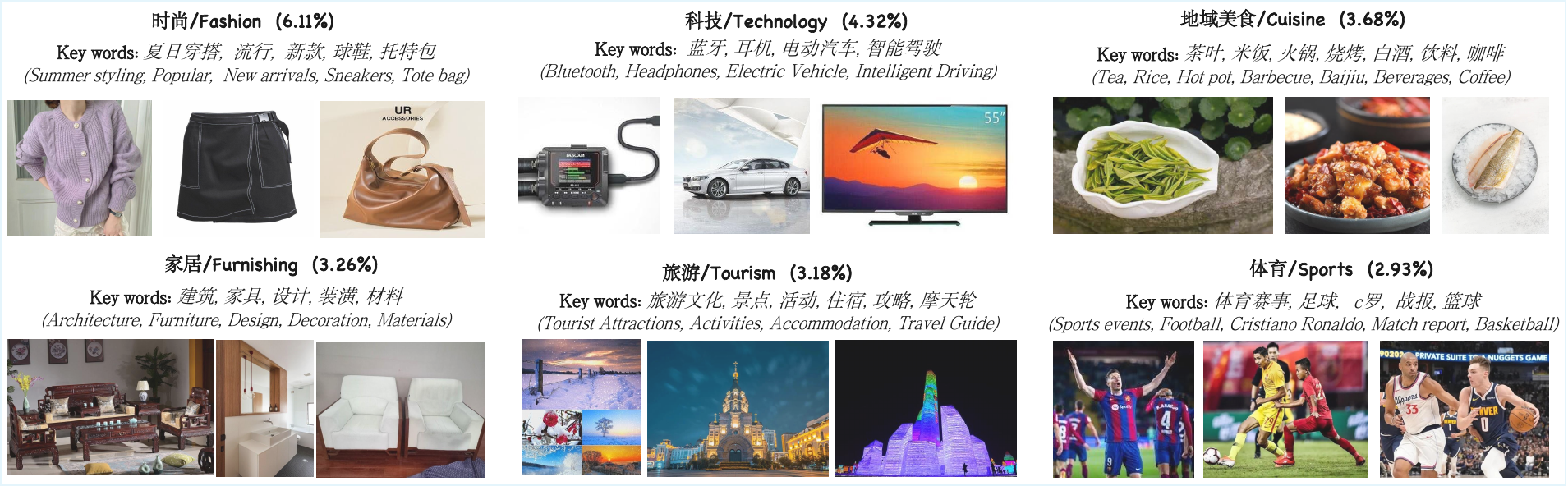}
    \vspace{-6mm}
    \caption{Visualization of popular topic in the \dsname\ dataset, generated via BERTopic~\cite{grootendorst2022bertopicneuraltopicmodeling} on 10M subset.}
    \vspace{-4mm}
    \label{fig:topic_example_main}
\end{figure*}

\noindent\textbf{Topic Modeling.} To further investigate the semantic diversity of \dsname, we implement a topic modeling pipeline based on BERTopic~\cite{grootendorst2022bertopicneuraltopicmodeling}. Specifically, we randomly sample 10M image-text pairs and extract text embeddings using Chinese-CLIP-L/14~\cite{chineseclip}. To address the challenges of high-dimensional clustering, we apply Uniform Manifold Approximation and Projection~\cite{unmap_paper} for dimensionality reduction. Subsequently, we use HDBSCAN~\cite{grootendorst2022bertopicneuraltopicmodeling} to identify distinct semantic clusters, setting the minimum cluster size to 1,000 to ensure cluster stability and reduce noise. We then utilize class-based TF-IDF to extract representative keywords for each topic. Fig.~\ref{fig:topic_example_main} visualizes six prevalent topics, including fashion, technology, cuisine, furnishing, tourism, and sports. These results indicate that \dsname\ encompasses a wide variety of real-world domains, providing a comprehensive foundation for large-scale vision-language representation learning. For a more detailed topic analysis, please refer to the Appendix~\ref{appendix: topic modeling}.

\section{Experiments and Results}

\subsection{Implementation Details}
To validate the effectiveness of the DanQing dataset, we continue pre-training the SigLIP2~\cite{siglip2} model for 2 epochs using 16 $\times$ A800 (80G) GPUs. We employ AdamW~\cite{adamw} as the optimizer, initializing it with a learning rate of 1e-5 and a weight decay of 0.1. The batch size is set to $768 \times 16$. The momentum parameters $\beta_1$ and $\beta_2$ are set to 0.9 and 0.98, respectively. A learning rate warmup strategy is applied during the first 1,000 iterations to ensure training stability. The input image size
is 256 $\times$ 256, and the input text sequence length is truncated
or padded to 64. To ensure a fair comparison in our experiments, we randomly select 100M samples from both the Zero and TaiSu datasets for training.

\begin{table}
\centering
\small
\setlength{\tabcolsep}{5.0pt} 
\renewcommand{\arraystretch}{1.0}
\resizebox{0.82\columnwidth}{!}{%
\begin{tikzpicture}
\node[inner sep=2.5pt] (tbl) {%
  {
  \begin{NiceTabular}{@{}lccccccccccccc@{}}
     \textbf{Dataset} & \rotatebox[origin=lb]{90}{\smash{\textbf{Caltech101}}~\cite{caltech101}} & \rotatebox[origin=lb]{90}{\smash{\textbf{CIFAR10}}~\cite{cifar100}} & \rotatebox[origin=lb]{90}{\smash{\textbf{Country211}}~\cite{country211}} & \rotatebox[origin=lb]{90}{\smash{\textbf{DTD}}~\cite{DTD}} & \rotatebox[origin=lb]{90}{\smash{\textbf{Food101}}~\cite{Food101}} & \rotatebox[origin=lb]{90}{\smash{\textbf{MNIST}}~\cite{MNIST}} & \rotatebox[origin=lb]{90}{\smash{\textbf{Flowers102}}~\cite{Flowers102}} & \rotatebox[origin=lb]{90}{\smash{\textbf{Pets}}~\cite{Pets}} & \rotatebox[origin=lb]{90}{\smash{\textbf{RESISC45}}~\cite{Resisc45}} & \rotatebox[origin=lb]{90}{\smash{\textbf{Cars}}~\cite{stanford_cars}} & \rotatebox[origin=lb]{90}{\smash{\textbf{Memes}}~\cite{kiela2020hateful}} & \rotatebox[origin=lb]{90}{\smash{\textbf{VOC2007}}~\cite{Voc2007}} & \textbf{Avg.} \\
    \midrule
    \multicolumn{14}{c}{\textbf{\textit{Model Architecture: SigLIP2-B/32@256}}} \\
    \midrule
    \textcolor{gray}{Baseline}~\cite{siglip2} & \textcolor{gray}{77.0} & \textcolor{gray}{85.1} & \textcolor{gray}{8.2} & \textcolor{gray}{35.9} & \textcolor{gray}{55.1} & \textcolor{gray}{81.9} & \textcolor{gray}{37.6} & \textcolor{gray}{61.9} & \textcolor{gray}{56.3} & \textcolor{gray}{76.3} & \textcolor{gray}{49.4} & \textcolor{gray}{69.0} & \textcolor{gray}{57.8}\\
    Wukong~\cite{wukong} & 78.6 & 91.7 & 9.5 & 42.6 & 61.2 & 83.0 & \underline{61.4} & 71.3 & 58.1 & \underline{75.1} & 53.8 & 75.6 & 63.5\\
    Zero$^{*}$~\cite{ccmb} & \underline{79.3} & \underline{92.2} & \textbf{10.8} & \underline{45.1} & \underline{64.7} & \textbf{86.3} & \textbf{63.2} & \underline{76.7} & \underline{58.9} & 74.5 & 49.6 & \textbf{77.3} & \underline{64.9}\\
    TaiSu$^{*}$~\cite{taisu} & 78.5 & 90.9 & 5.7 & 43.5 & 53.6 & \underline{83.5} & 52.4 & 62.9 & 53.3 & 58.9 & \underline{54.0} & \textbf{77.3} & 59.5\\
    \rowcolor{rowcolor_blue} 
    \dsname& \textbf{79.7} & \textbf{93.0} & \underline{9.9} & \textbf{46.4} & \textbf{66.6} & 83.4 & 58.5 & \textbf{78.7} & \textbf{61.4} & \textbf{76.0} & \textbf{54.4} & \underline{77.1} & \textbf{65.4}\\
    \midrule
    \multicolumn{14}{c}{\textbf{\textit{Model Architecture: SigLIP2-B/16@256}}} \\
    \midrule
    \textcolor{gray}{Baseline}~\cite{siglip2} & \textcolor{gray}{77.3} & \textcolor{gray}{85.4} & \textcolor{gray}{10.7} & \textcolor{gray}{35.3} & \textcolor{gray}{60.8} & \textcolor{gray}{83.9} & \textcolor{gray}{38.1} & \textcolor{gray}{65.0} & \textcolor{gray}{59.8} & \textcolor{gray}{81.0} & \textcolor{gray}{51.0} & \textcolor{gray}{71.0} & \textcolor{gray}{59.9}\\
    
    Wukong~\cite{wukong} & 78.4 & 90.3 & 12.7 & 44.8 & 68.7 & 81.5 & \underline{63.6} & 76.0 & 59.0 & \underline{80.8} & \textbf{55.0} & 78.4 & 65.8 \\
    Zero$^{*}$~\cite{ccmb} & \underline{79.5} & \underline{91.3} & \textbf{13.9} & \underline{45.6} & \underline{70.5} & \textbf{84.6} & \textbf{65.5} & \underline{78.9} & \underline{60.6} & 80.2 & 51.0 & 79.0 & \underline{66.7} \\
    TaiSu$^{*}$~\cite{taisu} & 78.6 & 89.3 & 7.0 & 44.6 & 58.1 & 82.2 & 54.3 & 65.9 & 55.8 & 62.1 & \underline{54.2} & \underline{79.2} & 60.9 \\
    \rowcolor{rowcolor_blue} 
    \dsname & \textbf{80.2} & \textbf{93.2} & \underline{13.3} & \textbf{48.0} & \textbf{71.6} & \underline{83.5} & 62.2 & \textbf{81.8} & \textbf{63.5} & \textbf{81.7} & 53.2 & \textbf{79.6} & \textbf{67.7} \\
    \midrule
    \multicolumn{14}{c}{\textbf{\textit{Model Architecture: SigLIP2-L/16@256}}} \\
    \midrule
    \textcolor{gray}{Baseline}~\cite{siglip2} & \textcolor{gray}{76.7} & \textcolor{gray}{88.5} & \textcolor{gray}{15.9} & \textcolor{gray}{44.8} & \textcolor{gray}{72.0} & \textcolor{gray}{80.8} & \textcolor{gray}{49.7} & \textcolor{gray}{84.3} & \textcolor{gray}{63.9} & \textcolor{gray}{87.4} & \textcolor{gray}{49.2} & \textcolor{gray}{68.9} & \textcolor{gray}{65.2}\\
    Wukong~\cite{wukong} & 80.3 & 96.1 & 20.5 & 48.2 & 78.3 & 84.9 & 74.3 & 84.5 & \textbf{65.7} & 86.5 & 55.0 & 78.1 & 71.0 \\
    Zero$^{*}$~\cite{ccmb} & \underline{82.4} & \underline{96.3} & \textbf{22.6} & \underline{48.9} & \underline{81.9} & \textbf{86.4} & \textbf{75.9} & \underline{89.5} & \underline{65.3} & \underline{87.8} & 52.0 & \underline{79.7} & \underline{72.4} \\
    TaiSu$^{*}$~\cite{taisu} & 81.7 & 94.8 & 13.1 & 44.3 & 68.9 & 74.2 & 64.5 & 79.1 & 59.4 & 70.7 & \underline{55.6} & \underline{79.7} & 65.5 \\
    \rowcolor{rowcolor_blue} 
    \dsname & \textbf{83.5} & \textbf{96.7} & \underline{22.4} & \textbf{49.2} & \textbf{83.8} & \underline{85.2} & \underline{75.0} & \textbf{90.0} & 64.8 & \textbf{88.7} & \textbf{55.8} & \textbf{79.9} & \textbf{72.9} \\
    
    \end{NiceTabular}
  }%
};

\draw[line width=0.08pt, rounded corners=4pt]
  (tbl.south west) rectangle (tbl.north east);

\draw[line width=0.08pt, rounded corners=4pt, opacity=0.6]
  ([xshift=0.5pt,yshift=0.5pt]tbl.south west)
    rectangle
  ([xshift=0.5pt,yshift=0.5pt]tbl.north east);
\end{tikzpicture}
}
\vspace{-1mm}
\caption{Zero-shot image classification performance using models pretrained on different datasets. $^{*}$ indicates random sampling of 100 million image-text pairs. The best and second best scores are in \textbf{boldface} and \underline{underlined}.}
\vspace{-3mm}
\label{tab:zero_shot_image_classification}
\end{table}

\begin{table*}
\centering
\small
\setlength{\tabcolsep}{2pt}
\renewcommand{\arraystretch}{1} 

\resizebox{\textwidth}{!}{
\begin{tikzpicture}
\node[inner sep=2.5pt] (tbl) {
  {
  \begin{NiceTabular}{@{}lccccccccccccccccccc@{}}
    & \multicolumn{6}{c}{\textbf{Flickr30K-CN}~\cite{flickr30kcn}} & \multicolumn{6}{c}{\textbf{MSCOCO-CN}~\cite{cococn}} & \multicolumn{6}{c}{\textbf{MUGE}~\cite{m6}}  & \\
    & \multicolumn{3}{c}{\textbf{Text to Image}} & \multicolumn{3}{c}{\textbf{Image to Text}} & \multicolumn{3}{c}{\textbf{Text to Image}} & \multicolumn{3}{c}{\textbf{Image to Text}} & \multicolumn{3}{c}{\textbf{Text to Image}} & \multicolumn{3}{c}{\textbf{Image to Text}} \\
    \cmidrule(lr){2-7} \cmidrule(lr){8-13}  \cmidrule(lr){14-19}
     \textbf{Dataset} & \makecell[c]{\textbf{R@1}} & \makecell[c]{\textbf{R@5}} & \makecell[c]{\textbf{R@10}} & \makecell[c]{\textbf{R@1}} & \makecell[c]{\textbf{R@5}} & \makecell[c]{\textbf{R@10}} & \makecell[c]{\textbf{R@1}} & \makecell[c]{\textbf{R@5}} & \makecell[c]{\textbf{R@10}} & \makecell[c]{\textbf{R@1}} & \makecell[c]{\textbf{R@5}} & \makecell[c]{\textbf{R@10}} & \makecell[c]{\textbf{R@1}} & \makecell[c]{\textbf{R@5}} & \makecell[c]{\textbf{R@10}} & \makecell[c]{\textbf{R@1}} & \makecell[c]{\textbf{R@5}} & \makecell[c]{\textbf{R@10}} & \textbf{Avg.} \\
    \midrule
    \multicolumn{20}{c}{\textbf{\textit{Model Architecture: SigLIP2-B/32@256}}} \\
    \midrule
    \textcolor{gray}{Baseline}~\cite{siglip2}& \textcolor{gray}{45.4} & \textcolor{gray}{71.2} & \textcolor{gray}{80.6} & \textcolor{gray}{67.7} & \textcolor{gray}{88.9} & \textcolor{gray}{94.6} & \textcolor{gray}{49.6} & \textcolor{gray}{77.2} & \textcolor{gray}{87.8} & \textcolor{gray}{51.8} & \textcolor{gray}{81.1} & \textcolor{gray}{90.9} & \textcolor{gray}{38.3} & \textcolor{gray}{61.7} & \textcolor{gray}{69.9} & \textcolor{gray}{35.3} & \textcolor{gray}{60.7} & \textcolor{gray}{69.8} & \textcolor{gray}{67.9}\\    
    Wukong~\cite{wukong} & 49.8 & 75.8 & 83.7 & 68.2 & 89.8 & 95.6 & 54.4 & 81.6 & 90.7 & 56.5 & 84.8 & 83.2 & \underline{55.1} & 77.9 & \textbf{85.1} & 44.0 & \underline{71.2} & 80.1 & 74.3\\ 
    Zero$^*$~\cite{ccmb} & 49.5 & 76.5 & 84.4 & 68.7 & 90.5 & 95.1 & 53.9 & 84.0 & 92.1 & 56.9 & 84.6 & 93.3 & 54.5 & 77.7 & \underline{84.9} & 42.1 & 69.4 & 78.5 & 74.3\\
    TaiSu$^*$~\cite{taisu} & \textbf{60.5} & \textbf{84.2} & \textbf{90.3} & \textbf{77.8} & \textbf{94.4} & \textbf{97.2} & \textbf{65.7} & \textbf{90.7} & \textbf{96.0} & \textbf{65.5} & \textbf{88.9} & \underline{94.5} & \textbf{56.2} & \textbf{78.2} & 84.7 & \underline{44.1} & \underline{71.2} & \underline{80.3} & \textbf{78.9} \\
    \rowcolor{rowcolor_blue} 
      \dsname & \underline{54.2} & \underline{79.0} & \underline{86.6} & \underline{73.0} & \underline{92.2} & \underline{96.3} & \underline{60.1} & \underline{84.5} & \underline{93.8} & \underline{61.0} & \underline{88.3} & \textbf{96.3} & 54.8 & \underline{78.1} & \underline{84.9} & \textbf{45.3} & \textbf{72.1} & \textbf{80.7} & \underline{76.7} \\
    \midrule
    \multicolumn{20}{c}{\textbf{\textit{Model Architecture: SigLIP2-B/16@256}}} \\
    \midrule
    \textcolor{gray}{Baseline~\cite{siglip2}} & \textcolor{gray}{51.3} & \textcolor{gray}{76.6} & \textcolor{gray}{84.7} & \textcolor{gray}{73.5} & \textcolor{gray}{93.3} & \textcolor{gray}{96.7} & \textcolor{gray}{51.9} & \textcolor{gray}{79.7} & \textcolor{gray}{89.6} & \textcolor{gray}{54.7} & \textcolor{gray}{82.5} & \textcolor{gray}{91.9} & \textcolor{gray}{41.6} & \textcolor{gray}{64.6} & \textcolor{gray}{73.4} & \textcolor{gray}{38.9} & \textcolor{gray}{64.3} & \textcolor{gray}{73.5} & \textcolor{gray}{71.3} \\
    Wukong~\cite{wukong} & 56.5 & 81.8 & 88.4 & 74.8 & 94.2 & 97.8 & 57.5 & 83.3 & 92.0 & 61.0 & 86.0 & 93.7 & 60.1 & \textbf{81.7} & \textbf{87.7} & \underline{48.8} & \underline{75.3} & \underline{83.2} & 78.0\\
    Zero$^*$~\cite{ccmb} & 58.2 & 83.7 & 90.4 & 74.9 & 93.4 & 96.9 & 58.7 & 86.0 & \underline{94.4} & 60.0 & 84.8 & 93.1 & 59.6 & 80.8 & 86.8 & 46.2 & 72.9 & 81.3 & 77.9\\
    TaiSu$^*$~\cite{taisu} & \textbf{68.2} & \textbf{89.0} & \textbf{93.9} & \textbf{83.8} & \textbf{97.2} & \textbf{99.4} & \textbf{68.8} & \textbf{93.0} & \textbf{97.1} & \textbf{67.1} & \textbf{90.1} & \underline{95.9} & \underline{60.3} & 81.0 & 86.8 & 48.4 & 74.9 & 83.0 & \textbf{82.1} \\
    \rowcolor{rowcolor_blue} 
    \dsname & \underline{61.1} & \underline{84.9} & \underline{90.9} & \underline{80.6} & \underline{95.0} & \underline{97.9} & \underline{62.3} & \underline{86.6} & \underline{94.4} & \underline{64.7} & \underline{88.5} & \textbf{96.1} & \textbf{60.4} & \underline{81.3} & \underline{87.3} & \textbf{50.3} & \textbf{76.3} & \textbf{83.9} & \underline{80.1} \\
    \midrule
    \multicolumn{20}{c}{\textbf{\textit{Model Architecture: SigLIP2-L/16@256}}} \\
    \midrule
    \textcolor{gray}{Baseline~\cite{siglip2}} & \textcolor{gray}{53.5} & \textcolor{gray}{78.1} & \textcolor{gray}{85.5} & \textcolor{gray}{79.6} & \textcolor{gray}{95.7} & \textcolor{gray}{98.3} & \textcolor{gray}{51.7} & \textcolor{gray}{79.9} & \textcolor{gray}{89.0} & \textcolor{gray}{55.4} & \textcolor{gray}{81.9} & \textcolor{gray}{90.5} & \textcolor{gray}{50.2} & \textcolor{gray}{71.1} & \textcolor{gray}{78.5} & \textcolor{gray}{45.6} & \textcolor{gray}{70.4} & \textcolor{gray}{78.5} & \textcolor{gray}{74.1}\\
    Wukong~\cite{wukong} & 62.8 & 86.2 & 91.5 & 81.7 & \underline{96.2} & 98.5 & 61.0 & 85.9 & 93.5 & 62.9 & 88.7 & 95.1 & \underline{66.6} & 84.6 & \underline{90.1} & \underline{55.8} & \underline{80.7} & \underline{87.4} & 81.6\\
    Zero$^*$~\cite{ccmb} & 64.3 & 87.9 & 93.4 & 78.4 & 95.5 & 98.6 & 61.6 & 87.2 & 94.7 & 62.1 & 87.2 & 94.6 & 65.9 & \textbf{85.3} & \textbf{90.3} & 53.9 & 79.0 & 86.2 & 81.5\\
    TaiSu$^*$~\cite{taisu} & \textbf{72.6} & \textbf{91.7} & \textbf{95.8} & \textbf{87.8} & \textbf{98.7} & \textbf{99.7} & \textbf{71.4} & \textbf{92.6} & \textbf{97.3} & \textbf{69.2} & \underline{91.6} & \underline{96.8} & 66.0 & \underline{85.0} & 90.0 & 55.1 & 80.1 & 86.7 & \textbf{84.9} \\
    \rowcolor{rowcolor_blue} 
    \dsname & \underline{70.2} & \underline{90.3} & \underline{94.7} & \underline{86.3} & \textbf{98.7} & \underline{99.6} & \underline{65.9} & \underline{90.5} & \underline{95.4} & \underline{68.0} & \textbf{92.6} & \textbf{97.4} & \textbf{67.5} & 84.9 & \underline{90.1} & \textbf{56.8} & \underline{81.2} & \textbf{87.5} & \underline{84.3} \\
\end{NiceTabular}%
  }%
};

\draw[line width=0.08pt, rounded corners=4pt]
  (tbl.south west) rectangle (tbl.north east);

\draw[line width=0.08pt, rounded corners=4pt, opacity=0.6]
  ([xshift=0.5pt,yshift=0.5pt]tbl.south west)
    rectangle
  ([xshift=0.5pt,yshift=0.5pt]tbl.north east);
\end{tikzpicture}
}
\vspace{-1mm}
\caption{Cross-modal retrieval performance on short-caption datasets for models pretrained on various large-scale Chinese image-text datasets. $^{*}$ indicates random sampling of 100 million image-text pairs. The best and second-best results are highlighted in \textbf{bold} and \underline{underlined}, respectively.}
\vspace{-7mm}
\label{tab:cross_modal_retrieval_short}
\end{table*}

\subsection{Main Results}
\noindent\textbf{Zero-shot Classification.} 
As presented in Tab.~\ref{tab:zero_shot_image_classification}, we perform continual pre-training on three SigLIP2 backbone models (B/32, B/16, and L/16) using the Wukong, Zero, TaiSu, and DanQing datasets. Continual pre-training with these datasets significantly improves model performance, with DanQing yielding the most notable gains. Specifically, DanQing enhances performance by 7.6\%, 7.8\%, and 7.7\% on B/32, B/16, and L/16, respectively. Additionally, compared to the Wukong dataset, DanQing achieves a 1.9\% performance improvement across all three backbone models. Similarly, compared to the Zero dataset, DanQing provides average performance improvements of 0.5\%, 1.0\%, and 0.5\% on B/32, B/16, and L/16, respectively. These results highlight the high quality of the DanQing dataset and its effectiveness in Chinese image-text contrastive learning tasks.

\noindent\textbf{Cross-Modal Retrieval.} 
To further validate the effectiveness of the DanQing dataset, we conduct comparisons on cross-modal retrieval tasks. As shown in Tab.~\ref{tab:cross_modal_retrieval_short}, on the Flickr30K-CN, MSCOCO-CN, and MUGE datasets, DanQing achieves average retrieval performance improvements of 2.4\%\&2.4\%, 2.1\%\&2.2\%, and 2.7\%\&2.8\% over the Wukong and Zero datasets across three different backbone models. Notably, TaiSu achieves strong retrieval performance, particularly on Flickr30K-CN and MSCOCO-CN, largely because it augments web-crawled tags with concise synthetic captions generated by OFA-Large~\cite{wang2022ofaunifyingarchitecturestasks}. The close alignment between the distribution of these captions and the target benchmarks results in substantial performance gains.

\begin{wraptable}[]{r}{0.65\textwidth}
\centering
\small
\vspace{-3mm}
\begin{minipage}[c]{0.65\textwidth}
    \centering
    \setlength{\tabcolsep}{4pt}
    \renewcommand{\arraystretch}{1}
    \resizebox{\linewidth}{!}{
    \begin{tikzpicture}
    \node[inner sep=2.5pt] (tbl) {
    {
    \begin{NiceTabular}{@{}lccccccccccccc@{}}
    & \multicolumn{6}{c}{\textbf{DCI-CN}~\cite{dci_paper}} & \multicolumn{6}{c}{\textbf{DOCCI-CN}~\cite{docci_papper}}  \\
    & \multicolumn{3}{c}{\textbf{Text to Image}} & \multicolumn{3}{c}{\textbf{Image to Text}} & \multicolumn{3}{c}{\textbf{Text to Image}} & \multicolumn{3}{c}{\textbf{Image to Text}}  \\
    \cmidrule(lr){2-7} \cmidrule(lr){8-13} 
    \textbf{Dataset} & \makecell[c]{\textbf{R@1}} & \makecell[c]{\textbf{R@5}} & \makecell[c]{\textbf{R@10}} & \makecell[c]{\textbf{R@1}} & \makecell[c]{\textbf{R@5}} & \makecell[c]{\textbf{R@10}} & \makecell[c]{\textbf{R@1}} & \makecell[c]{\textbf{R@5}} & \makecell[c]{\textbf{R@10}} & \makecell[c]{\textbf{R@1}} & \makecell[c]{\textbf{R@5}} & \makecell[c]{\textbf{R@10}} &\textbf{Avg.} \\
    \midrule
    \multicolumn{14}{c}{\textbf{\textit{Model Architecture: SigLIP2-B/32@256}}} \\
    \midrule
    \textcolor{gray}{Baseline}~\cite{siglip2}& \textcolor{gray}{7.7} & \textcolor{gray}{18.2} & \textcolor{gray}{23.8} & \textcolor{gray}{8.7} & \textcolor{gray}{19.3} & \textcolor{gray}{25.3} & \textcolor{gray}{11.0} & \textcolor{gray}{26.5} & \textcolor{gray}{36.0} & \textcolor{gray}{14.3} & \textcolor{gray}{32.1} & \textcolor{gray}{41.6} & \textcolor{gray}{22.0} \\    
    Wukong~\cite{wukong} & 10.2 & 22.7 & 29.6 & 11.3 & 23.8 & 30.7 & 16.3 & 34.8 & 44.8 & 15.7 & 35.7 & 46.3 & 26.8  \\ 
    Zero$^*$~\cite{ccmb} & 10.9 & 24.4 & \underline{32.0} & 11.2 & 23.7 & 30.9 & \underline{17.1} & 35.8 & 45.8 & \underline{17.8} & 38.5 &  \underline{50.1} & 28.2\\
    TaiSu$^*$~\cite{taisu} & \underline{11.3} & \underline{24.0} & 31.5 & \underline{12.5} & \textbf{26.2} & \underline{33.1} & 16.8 & \underline{37.1} & \underline{47.4} & 16.6 & \underline{38.8} & 49.1  & \underline{28.7} \\
    \rowcolor{rowcolor_blue} 
    \dsname & \textbf{13.1} & \textbf{27.4} & \textbf{35.0} & \textbf{12.6} & \underline{26.1} & \textbf{33.7} & \textbf{19.8} & \textbf{42.0} & \textbf{52.8} & \textbf{18.7} & \textbf{40.4} & \textbf{51.5}  &\textbf{31.1} \\
    \midrule
    \multicolumn{14}{c}{\textbf{\textit{Model Architecture: SigLIP2-B/16@256}}} \\
    \midrule
    \textcolor{gray}{Baseline~\cite{siglip2}} & \textcolor{gray}{8.7} & \textcolor{gray}{19.6} & \textcolor{gray}{25.9} & \textcolor{gray}{10.4} & \textcolor{gray}{21.0} & \textcolor{gray}{26.8} & \textcolor{gray}{13.0} & \textcolor{gray}{30.2} & \textcolor{gray}{39.8}  & \textcolor{gray}{16.6} & \textcolor{gray}{35.8} & \textcolor{gray}{46.8} & \textcolor{gray}{24.6} \\
    Wukong~\cite{wukong} & 12.2 & 25.3 & 32.6 & 12.8 & 25.8 & 32.6 & 17.7 & 39.1 & 49.4 & 18.0 & 38.9 &  50.1 & 29.5 \\
    Zero$^*$~\cite{ccmb} & 12.9 & 26.9 & \underline{34.8} & 12.9 & 25.6 & 33.0 & 18.8 & 39.5 & 49.5 & \underline{19.0} & 40.5 &  51.7 & 30.4\\
    TaiSu$^*$~\cite{taisu} & \underline{13.2} & \underline{27.1} & 34.7 & \underline{14.6} & \underline{28.6} & \underline{36.0} & \underline{19.1} & \underline{40.5} & \underline{51.5} & 18.9 & \underline{41.1} & \underline{51.9}  & \underline{31.4} \\
    \rowcolor{rowcolor_blue} 
    \dsname & \textbf{15.3} & \textbf{30.7} & \textbf{38.4} & \textbf{15.0} & \textbf{29.3} & \textbf{36.9} & \textbf{23.6} & \textbf{47.3} & \textbf{57.8} & \textbf{22.3} & \textbf{44.9} & \textbf{56.6} & \textbf{34.8} \\
    \midrule
    \multicolumn{14}{c}{\textbf{\textit{Model Architecture: SigLIP2-L/16@256}}} \\
    \midrule
    \textcolor{gray}{Baseline~\cite{siglip2}} & \textcolor{gray}{16.7} & \textcolor{gray}{30.9} & \textcolor{gray}{37.5} & \textcolor{gray}{16.3} & \textcolor{gray}{30.6} & \textcolor{gray}{38.0} & \textcolor{gray}{29.3} & \textcolor{gray}{53.5} & \textcolor{gray}{64.3} & \textcolor{gray}{29.0} & \textcolor{gray}{54.4} & \textcolor{gray}{64.9} & \textcolor{gray}{38.8} \\
    Wukong~\cite{wukong} & 23.1 & 41.0 & 48.6 & 21.8 & 38.3 & 46.0 & 37.1 & 64.2 & 74.2 & 33.3 & 59.2 & 69.8 & 46.4 \\
    Zero$^*$~\cite{ccmb} & 24.8 & 43.0 & 51.4 & 24.5 & 41.6 & 49.9 & 37.6 & 66.4 & 75.8 & \underline{38.4} & \underline{67.4} & \underline{77.8} & 49.9 \\
    TaiSu$^*$~\cite{taisu} & \underline{26.6} & \underline{44.4} & \underline{52.2} & \underline{26.0} & \underline{43.2} & \underline{51.1} & \underline{41.1} & \underline{67.3} & \underline{76.8} & 37.1 & 63.2 & 73.2 & \underline{50.2} \\
    \rowcolor{rowcolor_blue} 
    \dsname & \textbf{31.3} & \textbf{50.7} & \textbf{58.4} & \textbf{30.5} & \textbf{49.9} & \textbf{58.2} & \textbf{48.7} & \textbf{76.4} & \textbf{84.5} & \textbf{44.8} & \textbf{72.2} & \textbf{81.5} & \textbf{57.3} \\
\end{NiceTabular}%
  }%
};

\draw[line width=0.08pt, rounded corners=4pt]
  (tbl.south west) rectangle (tbl.north east);

\draw[line width=0.08pt, rounded corners=4pt, opacity=0.6]
  ([xshift=0.5pt,yshift=0.5pt]tbl.south west)
    rectangle
  ([xshift=0.5pt,yshift=0.5pt]tbl.north east);
\end{tikzpicture}
}
\vspace{-1mm}
\caption{Cross-modal retrieval performance on long-caption datasets for models pretrained on various datasets.}
\vspace{-3mm}
\label{tab:long_retrieval}
\end{minipage}
\end{wraptable}
In addition, inspired by recent work~\cite{fgclip2}, we further evaluate long caption-based cross-modal retrieval using the DCI-CN and DOCCI-CN datasets as shown in Tab.~\ref{tab:long_retrieval}. We surprisingly observe that, under the same pre-training context length (up to 64 tokens), the \dsname\ dataset demonstrates superior long caption-based cross-modal retrieval performance compared to other datasets. With the SigLIP2-L/16@256 model, \dsname\ achieves average performance improvements of 12.8\%, 9.0\%, and 8.9\% over the Wukong, Zero, and TaiSu datasets, respectively. This improvement is mainly because \dsname\ exhibits higher semantic density and a greater proportion of high-quality texts~(Fig.~\ref{fig:text information density}).

\noindent\textbf{Chinese-Centric Large Multimodal Model Tasks.} 
We employ SigLIP2 variants that continue pretrained on different Chinese image–text datasets as vision encoders in Large Multimodal Models (LMMs) to evaluate their compatibility and utility for Chinese-centric multimodal reasoning. \begin{wraptable}[]{r}{0.58\textwidth}
\centering
\small
\begin{minipage}[c]{0.58\textwidth}
    \centering
    \setlength{\tabcolsep}{4pt}
    \renewcommand{\arraystretch}{1}
    \resizebox{\linewidth}{!}{%
    \begin{tikzpicture}
    \node[inner sep=2.5pt] (tbl) {%
    {
    \begin{NiceTabular}{lcccccc}
     & \multicolumn{2}{c}{\textbf{MMBench (Dev)}} & \multirow{2}{*}{\textbf{MME-RW}} & \multirow{2}{*}{\textbf{CMMMU}} & \multirow{2}{*}{\textbf{OCRBench}} &  \\
    \cmidrule(lr){2-3}
     \textbf{Dataset}& \textbf{CN}~\cite{liu2024mmbench} & \textbf{EN}~\cite{liu2024mmbench} & \textbf{CN}~\cite{zhang2024mme} & \textbf{eval}~\cite{zhang2024cmmmu} & \textbf{V2}~\cite{fu2024ocrbench} &  \textbf{Avg.}\\
    \midrule
    \multicolumn{7}{c}{\textbf{\textit{Model Architecture: SigLIP2-L/16@256 + Qwen2-7B}}} \\
    \midrule
    \textcolor{gray}{Baseline}~\cite{siglip2} & \textcolor{gray}{72.9} & \textcolor{gray}{73.6} & \textcolor{gray}{43.1} & \textcolor{gray}{38.7} & \textcolor{gray}{15.0} & \textcolor{gray}{48.7} \\
    Wukong~\cite{wukong}   & 73.5 & \textbf{75.6} & \underline{43.8} & 39.7 & 15.0 & \underline{49.5} \\
    Zero$^*$~\cite{ccmb}     & 72.9 & 75.0 & 42.3 & 39.4 & \underline{15.7} & 49.1 \\
    TaiSu$^*$~\cite{taisu}    & \underline{73.5} & 75.3 & 42.8 & \textbf{39.8} & 15.1 & 49.3 \\
    \rowcolor{rowcolor_blue} \dsname  & \textbf{74.0} & \underline{75.3} & \textbf{45.4} & \underline{39.7} & \textbf{16.0} & \textbf{50.1} \\
\end{NiceTabular}
  }%
};

\draw[line width=0.08pt, rounded corners=4pt]
  (tbl.south west) rectangle (tbl.north east);

\draw[line width=0.08pt, rounded corners=4pt, opacity=0.6]
  ([xshift=0.5pt,yshift=0.5pt]tbl.south west)
    rectangle
  ([xshift=0.5pt,yshift=0.5pt]tbl.north east);
\end{tikzpicture}
}
\vspace{-2mm}
\caption{Performance of LLaVA-NeXT-style models on Chinese-centric LMM downstream benchmarks, utilizing vision encoders pretrained on various datasets.}
\vspace{-4mm}
\label{tab:MLLM_downstream}
\end{minipage}
\end{wraptable}Specifically, we strictly adhere to the LLaVA-NeXT~\cite{liu2024llavanext} training pipeline and data configuration, varying only the vision encoder~(SigLIP2-L/16) to isolate the effect of pretraining data on downstream multimodal capabilities. Results in Tab.~\ref{tab:MLLM_downstream} show that \dsname\ surpasses existing datasets, achieving a new state-of-the-art average score of 50.1\% (\textit{vs.} 49.5\%). These performance improvements demonstrate the higher data quality of the \dsname\ and highlight its potential for Chinese-centric tasks in LMM architectures.

\section{Analysis}

\subsection{Text Quality Analysis}
\begin{figure}[t!]
  \centering
  \begin{subfigure}[b]{0.49\textwidth}
    \centering
    \includegraphics[width=\textwidth]{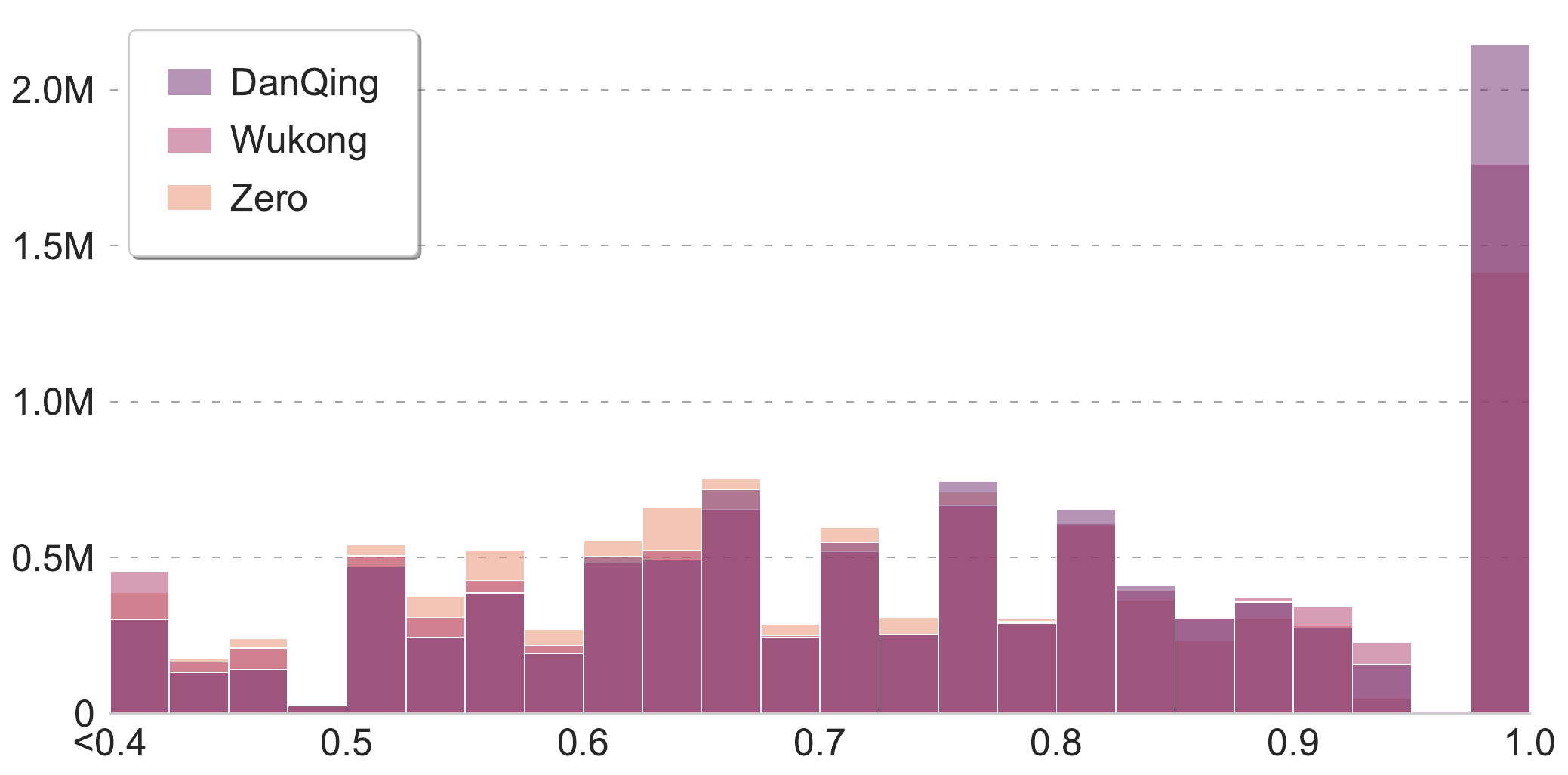}
    \vspace{-6mm}
    \caption{Text content word density}
    \label{fig:text word}
  \end{subfigure}
  \hfill
  \begin{subfigure}[b]{0.49\textwidth}
    \centering
    \includegraphics[width=\textwidth]{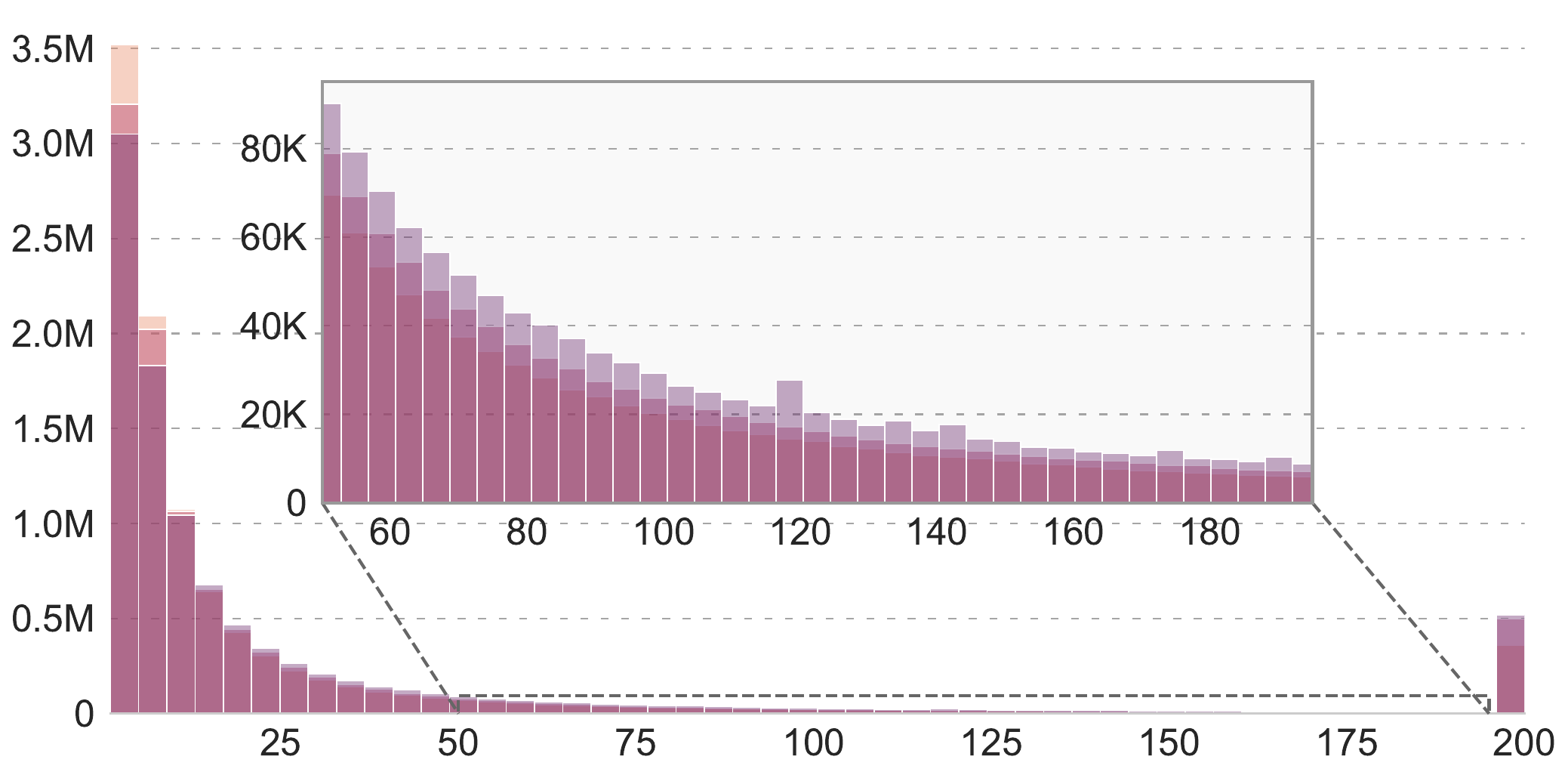}
    \vspace{-6mm}
    \caption{Text perplexit}
    \label{fig:text ppl}
  \end{subfigure}
  \hfill
  \vspace{-2mm}
  \caption{Text Quality Analysis. Randomly sample 10M subsets from \dsname, Wukong, and Zero, then compare their content word density and perplexity.}
  \vspace{-3mm}
  \label{fig:text information density}
\end{figure}
We further explore the text quality of \dsname\ through two metrics: semantic word density and perplexity (PPL), as illustrated in Fig.~\ref{fig:text information density}. Specifically, we randomly sampled 10M texts from \dsname, Wukong, and Zero for comparison. We use the jieba~\footnote{https://github.com/fxsjy/jieba} toolkit to identify semantic words (\textit{i.e.}, nouns, verbs, and adjectives) in each sentence and compute their proportions as a measure of semantic density. As illustrated in Fig.~\ref{fig:text word}, \dsname\ exhibits significantly higher semantic density than the other datasets, which enables the model to acquire more effective semantic information. Additionally, we compute sentence-level perplexity using a pre-trained Chinese BERT model~\cite{cnbert}. As shown in Fig.~\ref{fig:text ppl}, the number of samples in DanQing with PPL scores within the [50, 200] range is substantially higher than that of the other datasets. This range suggests an optimal level of linguistic complexity (neither overly simplistic nor incoherent), thereby highlighting the superior quality of our dataset for vision-language pre-training.

\subsection{Scaling Capability}
\begin{figure}[t!]
  \centering
  \begin{subfigure}[b]{0.48\textwidth}
    \centering
    \includegraphics[width=\textwidth]{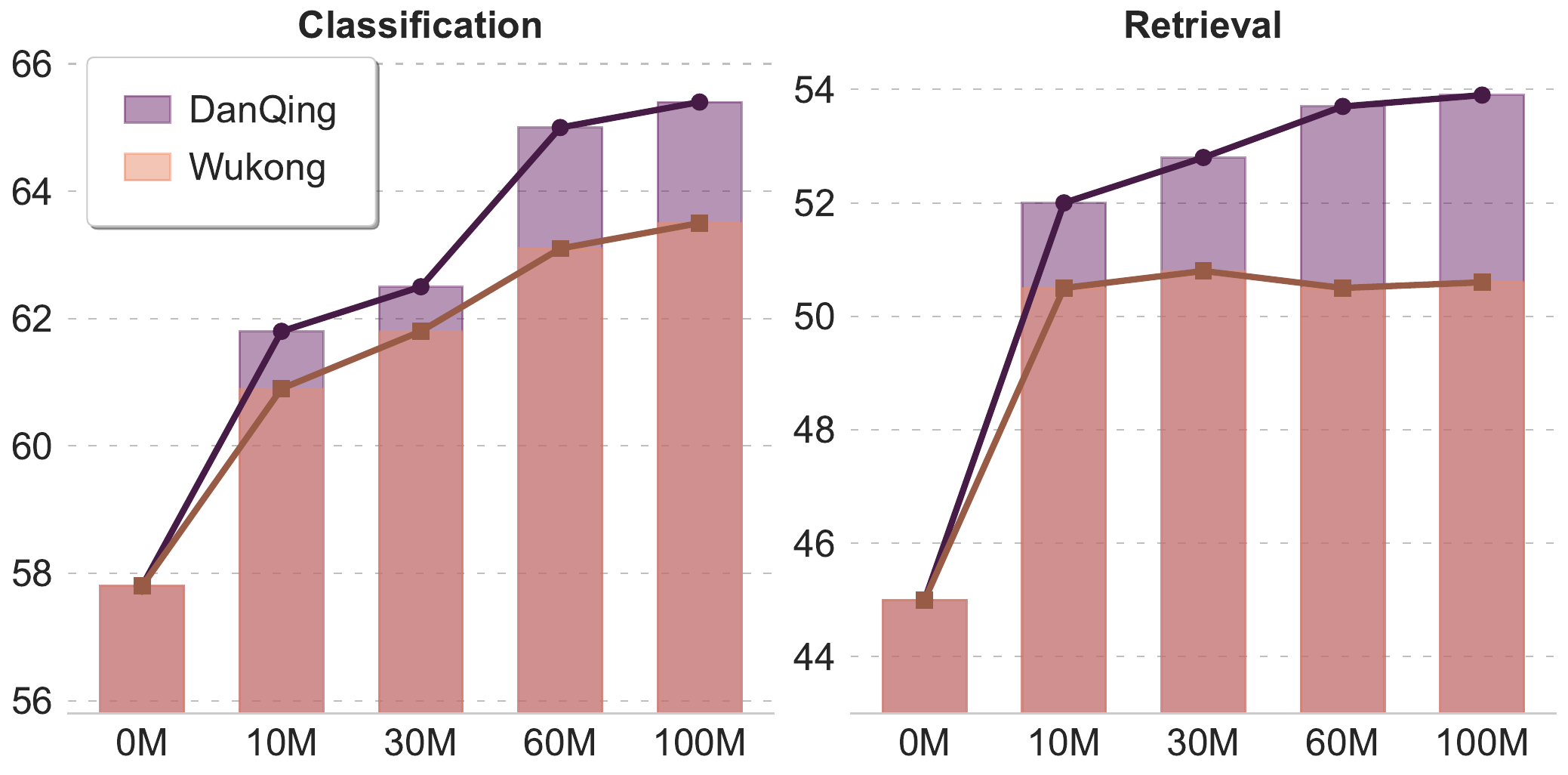}
    \vspace{-6mm}
    \caption{Data Scaling on SigLIP2-B/32@256}
    \label{fig:data scaling}
  \end{subfigure}
  \hfill
  \begin{subfigure}[b]{0.48\textwidth}
    \centering
    \includegraphics[width=\textwidth]{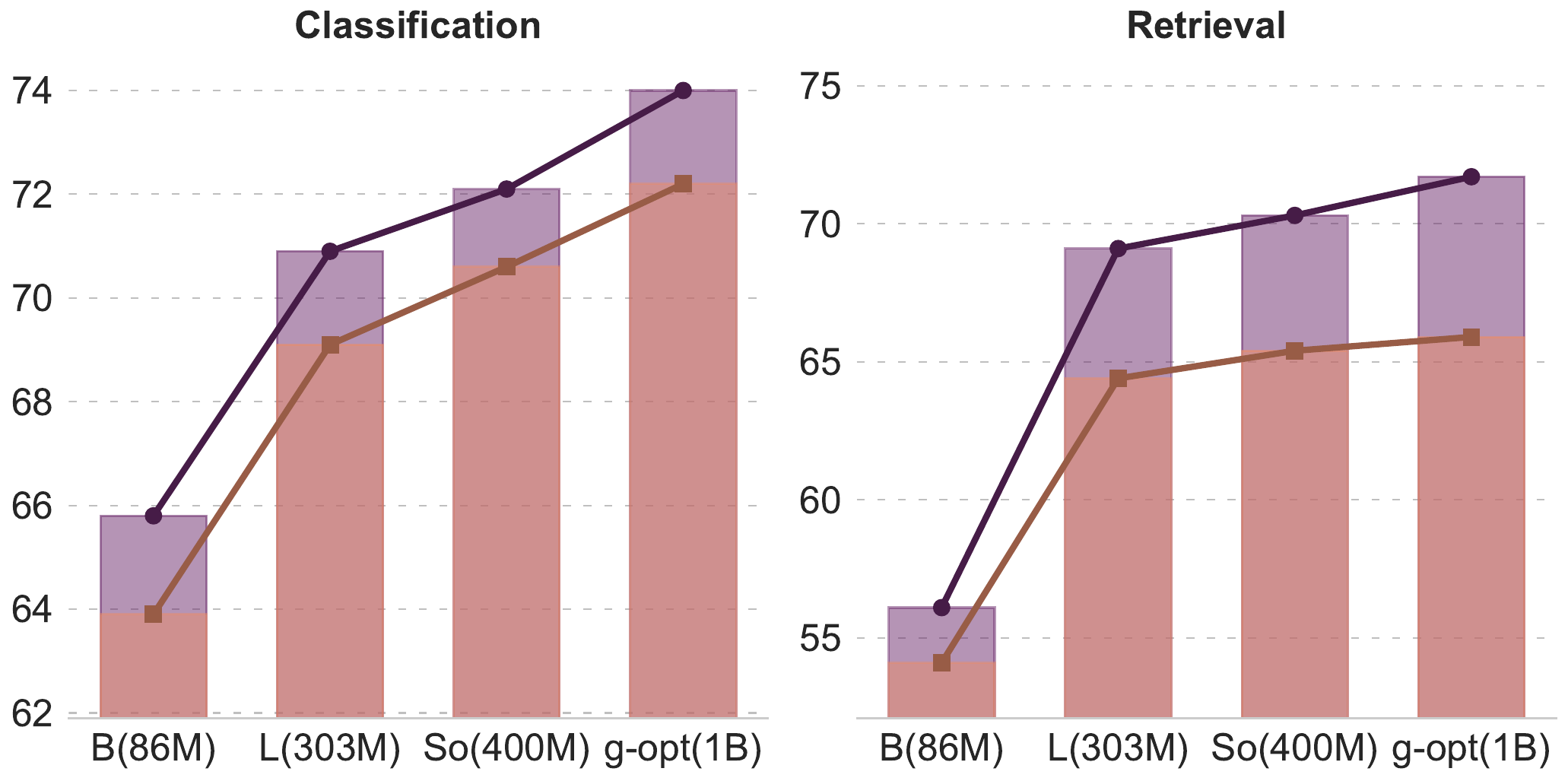}
    \vspace{-6mm}
    \caption{Model Scaling on 30M subset}
    \label{fig:model scaling}
  \end{subfigure}
  \hfill
  \vspace{-3mm}
  \caption{Data scaling and model scaling capability comparison between \dsname\ and Wukong.}
  \vspace{-5mm}
  \label{fig:scaling ability}
\end{figure}

Scaling capability determines the upper bound of vision-language pretraining models. To this end, we compare the data and model scaling capabilities of the proposed \dsname\ dataset with those of the widely used Wukong dataset, and report the average performance on zero-shot classification and retrieval (long\&short caption) tasks in Fig.~\ref{fig:scaling ability}.

\noindent\textbf{Data Scaling.} To evaluate the scalability of our proposed \dsname, we pretrain SigLIP2-B/32 for two epochs on varying data scales (10M, 30M, 60M, and 100M) and compare the performance trajectories of \dsname\ and Wukong. As shown in Fig.~\ref{fig:data scaling}, \dsname\ consistently achieves significant performance gains over Wukong across all data scales, with the improvements becoming more pronounced as the scale increases. Notably, Wukong’s retrieval performance plateaus beyond 30M samples, whereas \dsname\ continues to improve steadily from 30M to 100M, indicating that our dataset provides more effective supervision for large-scale vision-language pretraining.

\noindent\textbf{Model Scaling.}
To further investigate model scaling, we conduct experiments as illustrated in Fig.~\ref{fig:model scaling}. Specifically, we utilize 30M subsets from both \dsname\ and Wukong to train SigLIP2 models across various scales, including Base~(86M), Large~(303M), So~(400M), and g-opt~(1B). The results show that \dsname\ consistently outperforms Wukong in both classification and retrieval tasks. Moreover, \dsname\ exhibits a steeper scaling curve, better leveraging increased model capacity for representation learning.

\subsection{Image Semantic Balance}
\begin{wrapfigure}{l}{0.45\linewidth} 
    \centering
    \vspace{-4mm}
    \includegraphics[width=\linewidth]{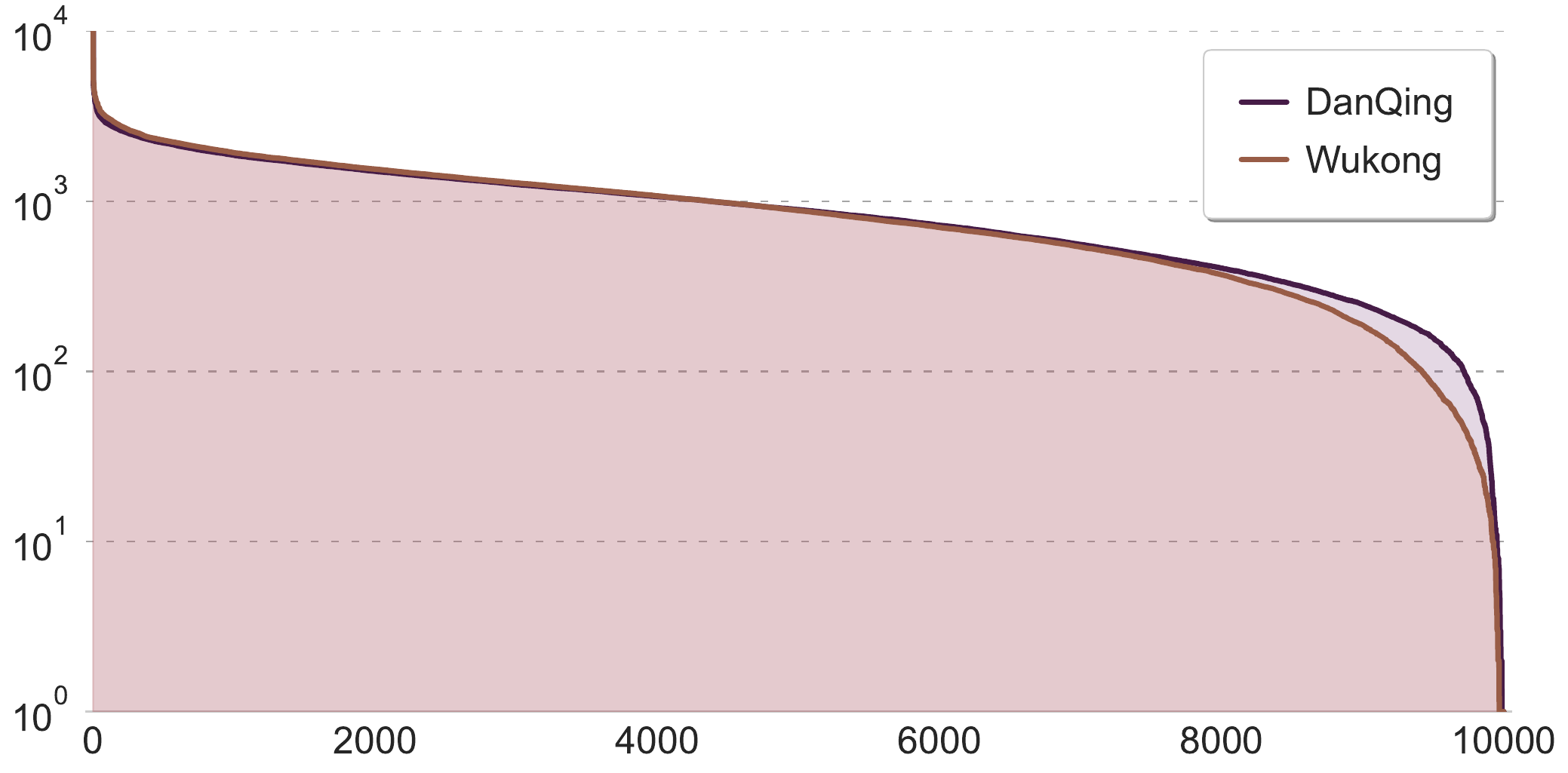}
    \vspace{-7mm}
    \caption{Clustering distribution of 10M subsets of \dsname\ and Wukong.}
    \vspace{-4mm}
    \label{fig:image semantic diversity}
\end{wrapfigure}

Fig.~\ref{fig:image semantic diversity} illustrates the semantic distribution of images in \dsname\ compared to the Wukong dataset. For quantitative analysis, we randomly sample 10 million images from each dataset and cluster them into 10k groups using the FAISS library~\cite{faiss}. We rank these clusters by the number of samples they contain. The results indicate that \dsname\ achieves a significantly more balanced and uniform semantic distribution than Wukong, effectively mitigating the long-tail effect. This increased uniformity suggests broader coverage of the visual manifold, which is essential for learning rare or long-tail concepts during vision-language pretraining.

\subsection{Image-Text Alignment}
\begin{wrapfigure}{r}{0.45\linewidth} 
    \centering
    \vspace{-4mm}
    \includegraphics[width=\linewidth]{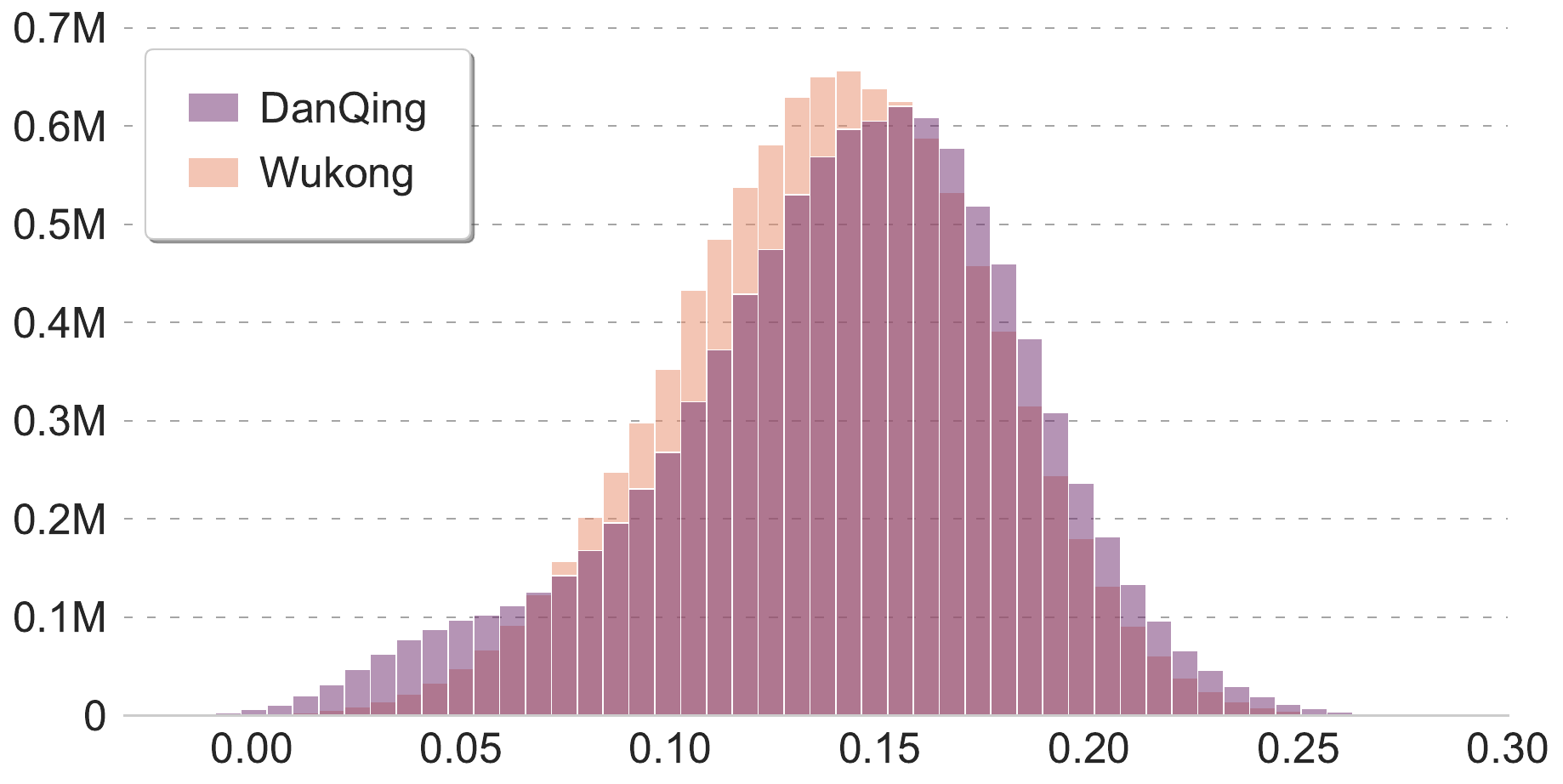}
    \vspace{-8mm}
    \caption{Similarity distributions for 10M subsets of \dsname\ and Wukong.}
    \vspace{-3mm}
    \label{fig:image-text alignment}
\end{wrapfigure}

In Fig.~\ref{fig:image-text alignment}, we illustrate the distribution of image-text similarity scores for 10M-sample subsets of \dsname\ and Wukong~\cite{wukong}. 
We employ the state-of-the-art Chinese retrieval model FG-CLIP2-L/16~\cite{fgclip2} to extract multimodal features and compute their cosine similarity. The results indicate that \dsname\ consistently achieves higher similarity scores than Wukong, with significantly more samples exceeding the 0.15 threshold. This demonstrates that \dsname\ provides stronger semantic consistency between images and texts. It is noteworthy that the DanQing dataset contains a significantly higher proportion of samples in the 0 to 0.05 similarity range compared to Wukong. This is primarily because DanQing comprises data from 2024 and 2025, which includes a substantial amount of novel semantic content (as shown in Fig.~\ref{fig:new concept understanding}). These findings help explain why models trained on \dsname\ demonstrate significant performance improvements in retrieval tasks, further highlighting the dataset's ability to enrich models with comprehensive semantic knowledge.

\subsection{New Concept Understanding}
\begin{figure}[t!]
  \centering
  \begin{subfigure}[b]{0.49\textwidth}
    \centering
    \includegraphics[width=\textwidth]{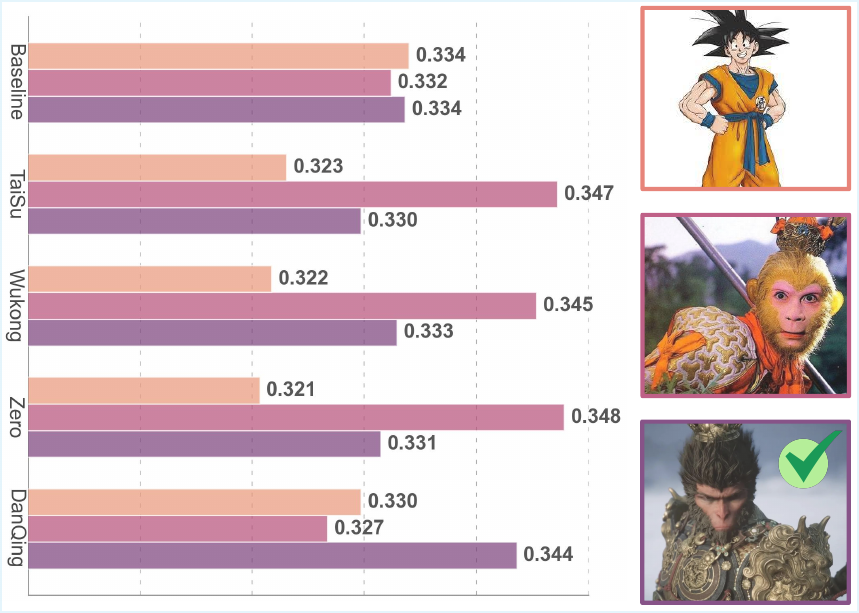}
    \vspace{-5mm}
    \caption{\begin{CJK*}{UTF8}{gkai}
		``黑神话:悟空''
	\end{CJK*}(\textcolor{gray}{Black Myth: Wukong)}}
    \label{fig:image resolution}
  \end{subfigure}
  \hfill
  \begin{subfigure}[b]{0.49\textwidth}
    \centering
    \includegraphics[width=\textwidth]{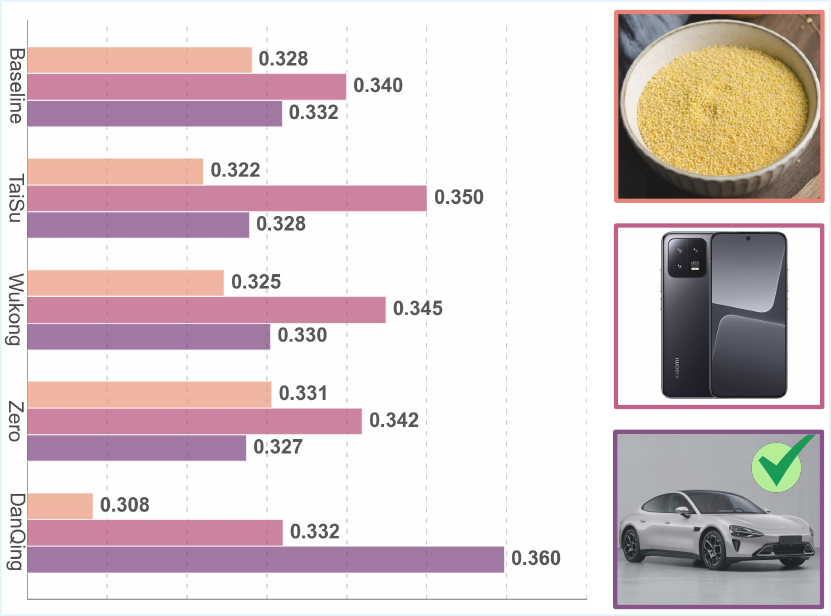}
    \vspace{-5mm}
    \caption{\begin{CJK*}{UTF8}{gkai}
		``小米SU7''
	\end{CJK*}\textcolor{gray}{(Xiaomi SU7)}}
    \label{fig:image2}
  \end{subfigure}
  \hfill
  \vspace{-2mm}
  \caption{New concept understanding capability comparison between DanQing with existing datasets. The scores represent the softmax-normalized values of the cosine similarities among three image-text pairs.}
  \vspace{-3mm}
  \label{fig:new concept understanding}
\end{figure}
In Fig.~\ref{fig:new concept understanding}, we evaluate the capability of SigLIP2-L/16@256 models pre-trained on different Chinese datasets to understand emergent concepts. Specifically, we select internet buzzwords that appear after 2024, such as \begin{CJK*}{UTF8}{gkai}``黑神话:悟空''\end{CJK*} (Black Myth: Wukong) and \begin{CJK*}{UTF8}{gkai}``小米SU7''\end{CJK*} (Xiaomi SU7). We pair these keywords with their corresponding ground-truth images, along with several semantically related distractors (\textit{e.g.}, traditional cartoon or TV adaptations of ``Wukong'', other Xiaomi products, and unrelated food items). By calculating the image-text similarity scores, we observe that the model trained on \dsname\ consistently assigns the highest confidence to the correct pairs. This superiority demonstrates that \dsname\ contains more up-to-date information, effectively enabling models to internalize contemporary knowledge and generalize to recent real-world concepts.
\section{Conclusion}

In this paper, we present \dsname, a large-scale Chinese image–text dataset comprising approximately 100M pairs that mitigates the scarcity of high-quality cross-modal resources for Chinese vision-language pretraining. We develop a rigorous curation pipeline to ensure quality, filtering raw web data and retaining about 10\% as high-quality pairs. Through continued pretraining of SigLIP2 models, \dsname consistently outperforms existing Chinese datasets across multiple downstream benchmarks. Comprehensive analysis further reveals that \dsname exhibits stronger scaling capability, improved understanding of novel concepts, higher textual quality, and more balanced visual semantics distribution. To facilitate and accelerate future research, we will open-source the \dsname\ dataset, providing a robust foundation for large-scale Chinese vision–language pretraining.

\clearpage
{
\bibliography{Danqing}

@String(IJCV  = {Int. J. Comput. Vis.})

@String(CVPR  = {IEEE Conf. Comput. Vis. Pattern Recog.})

@String(ICCV  = {Int. Conf. Comput. Vis.})

@String(ECCV  = {Eur. Conf. Comput. Vis.})

@String(NeurIPS = {Adv. Neural Inform. Process. Syst.})

@String(ICML  = {Int. Conf. Mach. Learn.})

@String(ICLR  = {Int. Conf. Learn. Represent.})

@String(CVPRW = {IEEE Conf. Comput. Vis. Pattern Recog. Worksh.})

@String(AAAI  = {AAAI})

@String(TMLR  = {Trans. Mach. Learn Res.})

@String(TMM   = {IEEE Trans. Multimedia})

@String(IJCV  = {IJCV})

@String(CVPR  = {CVPR})

@String(ICCV  = {ICCV})

@String(ECCV  = {ECCV})

@String(NeurIPS = {NeurIPS})

@String(ICML  = {ICML})

@String(ICLR  = {ICLR})

@String(CVPRW = {CVPRW})

@String(TMLR  = {TMLR})

@String(TMM   =	{IEEE TMM})

@inproceedings{cc3m,
  title = {Conceptual Captions: A Cleaned, Hypernymed, Image Alt-text Dataset For Automatic Image Captioning},
  author = {Sharma, Piyush and Ding, Nan and Goodman, Sebastian and Soricut, Radu},
  booktitle = {ACL},
  year = {2018},
}

@inproceedings{cc12m,
  title={Conceptual 12m: Pushing web-scale image-text pre-training to recognize long-tail visual concepts},
  author={Changpinyo, Soravit and Sharma, Piyush and Ding, Nan and Soricut, Radu},
  booktitle={CVPR},
  pages={3558--3568},
  year={2021}
}

@article{redcaps,
  title={Redcaps: Web-curated image-text data created by the people, for the people},
  author={Desai, Karan and Kaul, Gaurav and Aysola, Zubin and Johnson, Justin},
  journal={arXiv preprint arXiv:2111.11431},
  year={2021}
}

@inproceedings{wit,
  title={Wit: Wikipedia-based image text dataset for multimodal multilingual machine learning},
  author={Srinivasan, Krishna and Raman, Karthik and Chen, Jiecao and Bendersky, Michael and Najork, Marc},
  booktitle={SIGIR},
  pages={2443--2449},
  year={2021}
}

@article{yfcc100m,
  title={Yfcc100m: The new data in multimedia research},
  author={Thomee, Bart and Shamma, David A and Friedland, Gerald and Elizalde, Benjamin and Ni, Karl and Poland, Douglas and Borth, Damian and Li, Li-Jia},
  journal={Communications of the ACM},
  volume={59},
  number={2},
  pages={64--73},
  year={2016},
  publisher={ACM New York, NY, USA}
}

@misc{coyo700m,
  title         = {COYO-700M: Image-Text Pair Dataset},
  author        = {Byeon, Minwoo and Park, Beomhee and Kim, Haecheon and Lee, Sungjun and Baek, Woonhyuk and Kim, Saehoon},
  year          = {2022},
  howpublished  = {\url{https://github.com/kakaobrain/coyo-dataset}},
}

@article{laion400m,
  title={Laion-400m: Open dataset of clip-filtered 400 million image-text pairs},
  author={Schuhmann, Christoph and Vencu, Richard and Beaumont, Romain and Kaczmarczyk, Robert and Mullis, Clayton and Katta, Aarush and Coombes, Theo and Jitsev, Jenia and Komatsuzaki, Aran},
  journal={arXiv preprint arXiv:2111.02114},
  year={2021}
}

@article{laion5b,
  title={Laion-5b: An open large-scale dataset for training next generation image-text models},
  author={Schuhmann, Christoph and Beaumont, Romain and Vencu, Richard and Gordon, Cade and Wightman, Ross and Cherti, Mehdi and Coombes, Theo and Katta, Aarush and Mullis, Clayton and Wortsman, Mitchell and others},
  journal={Neurips},
  volume={35},
  pages={25278--25294},
  year={2022}
}

@article{datacomp,
  title={Datacomp: In search of the next generation of multimodal datasets},
  author={Gadre, Samir Yitzhak and Ilharco, Gabriel and Fang, Alex and Hayase, Jonathan and Smyrnis, Georgios and Nguyen, Thao and Marten, Ryan and Wortsman, Mitchell and Ghosh, Dhruba and Zhang, Jieyu and others},
  journal={Neurips},
  volume={36},
  pages={27092--27112},
  year={2023}
}

@inproceedings{product1m,
  title={Product1m: Towards weakly supervised instance-level product retrieval via cross-modal pretraining},
  author={Zhan, Xunlin and Wu, Yangxin and Dong, Xiao and Wei, Yunchao and Lu, Minlong and Zhang, Yichi and Xu, Hang and Liang, Xiaodan},
  booktitle={ICCV},
  pages={11782--11791},
  year={2021}
}

@inproceedings{m6,
  title={M6: Multi-modality-to-multi-modality multitask mega-transformer for unified pretraining},
  author={Lin, Junyang and Men, Rui and Yang, An and Zhou, Chang and Zhang, Yichang and Wang, Peng and Zhou, Jingren and Tang, Jie and Yang, Hongxia},
  booktitle={ACM SIGKDD},
  pages={3251--3261},
  year={2021}
}

@article{wudaomm,
  title={Wudaomm: A large-scale multi-modal dataset for pre-training models},
  author={Yuan, Sha and Zhao, Shuai and Leng, Jiahong and Xue, Zhao and Zhao, Hanyu and Liu, Peiyu and Gong, Zheng and Zhao, Wayne Xin and Li, Junyi and Tang, Jie},
  journal={arXiv preprint arXiv:2203.11480},
  year={2022}
}

@article{wukong,
  title={Wukong: A 100 million large-scale chinese cross-modal pre-training benchmark},
  author={Gu, Jiaxi and Meng, Xiaojun and Lu, Guansong and Hou, Lu and Minzhe, Niu and Liang, Xiaodan and Yao, Lewei and Huang, Runhui and Zhang, Wei and Jiang, Xin and others},
  journal={Neurips},
  volume={35},
  pages={26418--26431},
  year={2022}
}

@article{taisu,
  title={Taisu: A 166m large-scale high-quality dataset for chinese vision-language pre-training},
  author={Liu, Yulong and Zhu, Guibo and Zhu, Bin and Song, Qi and Ge, Guojing and Chen, Haoran and Qiao, GuanHui and Peng, Ru and Wu, Lingxiang and Wang, Jinqiao},
  journal={NIPS},
  volume={35},
  pages={16705--16717},
  year={2022}
}

@inproceedings{ccmb,
  title={Ccmb: A large-scale chinese cross-modal benchmark},
  author={Xie, Chunyu and Cai, Heng and Li, Jincheng and Kong, Fanjing and Wu, Xiaoyu and Song, Jianfei and Morimitsu, Henrique and Yao, Lin and Wang, Dexin and Zhang, Xiangzheng and others},
  booktitle={ACM MM},
  pages={4219--4227},
  year={2023}
}

@article{siglip2,
  title={Siglip 2: Multilingual vision-language encoders with improved semantic understanding, localization, and dense features},
  author={Tschannen, Michael and Gritsenko, Alexey and Wang, Xiao and Naeem, Muhammad Ferjad and Alabdulmohsin, Ibrahim and Parthasarathy, Nikhil and Evans, Talfan and Beyer, Lucas and Xia, Ye and Mustafa, Basil and others},
  journal={arXiv preprint arXiv:2502.14786},
  year={2025}
}

@article{joulin2016fasttext,
  title={FastText.zip: Compressing text classification models},
  author={Joulin, Armand and Grave, Edouard and Bojanowski, Piotr and Douze, Matthijs and J{\'e}gou, H{\'e}rve and Mikolov, Tomas},
  journal={arXiv preprint arXiv:1612.03651},
  year={2016}
}

@article{metaclip,
  title={Demystifying clip data},
  author={Xu, Hu and Xie, Saining and Tan, Xiaoqing Ellen and Huang, Po-Yao and Howes, Russell and Sharma, Vasu and Li, Shang-Wen and Ghosh, Gargi and Zettlemoyer, Luke and Feichtenhofer, Christoph},
  journal={arXiv preprint arXiv:2309.16671},
  year={2023}
}

@inproceedings{clip,
  title={Learning Transferable Visual Models From Natural Language Supervision}, 
  author={Alec Radford and Jong Wook Kim and Chris Hallacy and Aditya Ramesh and Gabriel Goh and Sandhini Agarwal and Girish Sastry and Amanda Askell and Pamela Mishkin and Jack Clark and Gretchen Krueger and Ilya Sutskever},
  year={2021},
  booktitle={ICML},
}

@inproceedings{gu2025realsyn,
  title={Realsyn: An effective and scalable multimodal interleaved document transformation paradigm},
  author={Gu, Tiancheng and Yang, Kaicheng and Zhang, Chaoyi and Xie, Yin and An, Xiang and Feng, Ziyong and Liu, Dongnan and Cai, Weidong and Deng, Jiankang},
  booktitle={ACM MM},
  pages={3487--3496},
  year={2025}
}

@misc{opencv_library,
  author = {OpenCV team},
  title = {OpenCV: Open Source Computer Vision Library},
  howpublished = {\url{https://github.com/opencv/opencv}},
  year = {2024},
}

@misc{OpenCC,
    author    = {BYVoid},
    title     = {{OpenCC: Open Chinese Convert}},
    year      = {2024},
    howpublished       = {https://github.com/BYVoid/OpenCC},
}

@software{PaddleHubPornDetection2021,
  author       = {{Baidu AI Studio Community}},
  title        = {{PaddleHub Pornographic Content Detection Model Tutorial}},
  year         = {2021},
  month        = {Jan},
  version      = {3.0},
  url          = {https://aistudio.baidu.com/projectdetail/1444248},
  note         = {Online tutorial demonstrating the use of PaddleHub's \texttt{porn\_detection\_lstm} model for text-based content moderation. Last updated on 2021-01-13.}
}

@software{PaddleHubPornDetectionCNN2021,
  author       = {{Baidu AI Studio Community}},
  title        = {{PaddleHub Pornographic Content Detection Model Tutorial}},
  year         = {2021},
  month        = {Jan},
  version      = {3.0},
  url          = {https://www.paddlepaddle.org.cn/hubdetail?name=porn_detection_cnn&en_category=TextCensorship},
  note         = {Online tutorial demonstrating the use of PaddleHub's \texttt{porn\_detection\_lstm} model for text-based content moderation. Last updated on 2021-01-13.}
}

@inproceedings{rwkvclip,
  title={RWKV-CLIP: A Robust Vision-Language Representation Learner},
  author={Gu, Tiancheng and Yang, Kaicheng and An, Xiang and Feng, Ziyong and Liu, Dongnan and Cai, Weidong and Deng, Jiankang},
  booktitle={EMNLP},
  pages={4799--4812},
  year={2024}
}

@inproceedings{longclip,
  title={Long-clip: Unlocking the long-text capability of clip},
  author={Zhang, Beichen and Zhang, Pan and Dong, Xiaoyi and Zang, Yuhang and Wang, Jiaqi},
  booktitle={ECCV},
  pages={310--325},
  year={2024},
  organization={Springer}
}

@article{adamw,
  title={Decoupled weight decay regularization},
  author={Loshchilov, Ilya and Hutter, Frank},
  journal={arXiv preprint arXiv:1711.05101},
  year={2017}
}

@article{dfn,
  title={Data filtering networks},
  author={Fang, Alex and Jose, Albin Madappally and Jain, Amit and Schmidt, Ludwig and Toshev, Alexander and Shankar, Vaishaal},
  journal={arXiv preprint arXiv:2309.17425},
  year={2023}
}

@article{metaclip2,
  title={Meta clip 2: A worldwide scaling recipe},
  author={Chuang, Yung-Sung and Li, Yang and Wang, Dong and Yeh, Ching-Feng and Lyu, Kehan and Raghavendra, Ramya and Glass, James and Huang, Lifei and Weston, Jason and Zettlemoyer, Luke and others},
  journal={arXiv preprint arXiv:2507.22062},
  year={2025}
}

@software{freepik2025nsfw,
    title={EVA-based Fast NSFW Image Classifier},
    author={Freepik Company S.L.},
    year={2025},
    publisher={Hugging Face},
    url = {https://huggingface.co/Freepik/nsfw_image_detector},
    organization = {Freepik Company S.L.}
}

@inproceedings{clipcid,
  title={Clip-cid: Efficient clip distillation via cluster-instance discrimination},
  author={Yang, Kaicheng and Gu, Tiancheng and An, Xiang and Jiang, Haiqiang and Dai, Xiangzi and Feng, Ziyong and Cai, Weidong and Deng, Jiankang},
  booktitle={AAAI},
  volume={39},
  number={20},
  pages={21974--21982},
  year={2025}
}

@inproceedings{alip,
  title={Alip: Adaptive language-image pre-training with synthetic caption},
  author={Yang, Kaicheng and Deng, Jiankang and An, Xiang and Li, Jiawei and Feng, Ziyong and Guo, Jia and Yang, Jing and Liu, Tongliang},
  booktitle={ICCV},
  pages={2922--2931},
  year={2023}
}

@article{chineseclip,
  title={Chinese clip: Contrastive vision-language pretraining in chinese},
  author={Yang, An and Pan, Junshu and Lin, Junyang and Men, Rui and Zhang, Yichang and Zhou, Jingren and Zhou, Chang},
  journal={arXiv preprint arXiv:2211.01335},
  year={2022}
}

@inproceedings{siglip,
  title={Sigmoid loss for language image pre-training},
  author={Zhai, Xiaohua and Mustafa, Basil and Kolesnikov, Alexander and Beyer, Lucas},
  booktitle={ICCV},
  pages={11975--11986},
  year={2023}
}

@inproceedings{slip,
  title={Slip: Self-supervision meets language-image pre-training},
  author={Mu, Norman and Kirillov, Alexander and Wagner, David and Xie, Saining},
  booktitle={ECCV},
  pages={529--544},
  year={2022},
  organization={Springer}
}

@article{mokady2021clipcap,
  title={ClipCap: CLIP Prefix for Image Captioning},
  author={Mokady, Ron and Hertz, Amir and Bermano, Amit H},
  journal={arXiv preprint arXiv:2111.09734},
  year={2021}
}

@inproceedings{li2022blip,
  title={BLIP: Bootstrapping Language-Image Pre-training for Unified Vision-Language Understanding and Generation},
  author={Li, Junnan and Li, Dongxu and Xiong, Caiming and Hoi, Steven},
  booktitle={ICML},
  year={2022}
}

@article{yu2022coca,
  title={CoCa: Contrastive Captioners are Image-Text Foundation Models},
  author={Yu, Jiahui and Wang, Zirui and Vasudevan, Vijay and Yeung, Legg and Seyedhosseini, Mojtaba and Wu, Yonghui},
  journal={TMLR},
  year={2022}
}

@article{fgclip2,
  title={FG-CLIP 2: A Bilingual Fine-grained Vision-Language Alignment Model},
  author={Xie, Chunyu and Wang, Bin and Kong, Fanjing and Li, Jincheng and Liang, Dawei and Ao, Ji and Leng, Dawei and Yin, Yuhui},
  journal={arXiv preprint arXiv:2510.10921},
  year={2025}
}

@inproceedings{gu2022vild,
  title={Zero-Shot Detection via Vision and Language Knowledge Distillation},
  author={Gu, Xiuye and Lin, Tsung-Yi and Kuo, Weicheng and Cui, Yin andFeature Pyramid Networks for Object Detection},
  booktitle={ICLR},
  year={2022}
}

@inproceedings{li2022glip,
  title={Grounded Language-Image Pre-training},
  author={Li, Liunian Harold and Zhang, Pengchuan and Zhang, Haotian and Yang, Jianwei and Li, Chunyuan and Zhong, Yiwu and Wang, Lijuan and Yuan, Lu and Zhang, Lei and Hwang, Jenq-Neng and others},
  booktitle={CVPR},
  year={2022}
}

@inproceedings{zhong2022regionclip,
  title={RegionCLIP: Region-based Language-Image Pretraining},
  author={Zhong, Yiwu and Yang, Jianwei and Zhang, Pengchuan and Li, Chunyuan and Codella, Noel and Li, Liunian Harold and Zhou, Luowei and Dai, Xiyang and Yuan, Lu and Li, Yin and others},
  booktitle={CVPR},
  year={2022}
}

@inproceedings{li2022lseg,
  title={Language-driven Semantic Segmentation},
  author={Li, Boyi and Weinberger, Kilian Q and Belongie, Serge and Koltun, Vladlen and Ranftl, Ren{\'e}},
  booktitle={ICLR},
  year={2022}
}

@inproceedings{rao2022denseclip,
  title={DenseCLIP: Language-Guided Dense Prediction with Context-Aware Prompting},
  author={Rao, Yongming and Zhao, Wenliang and Light, Guangyi and Zhou, Jiwen and Lu, Jiwen and others},
  booktitle={CVPR},
  year={2022}
}

@inproceedings{xu2022groupvit,
  title={GroupViT: Semantic Segmentation Emerges from Text Supervision},
  author={Xu, Jiarui and De Mello, Shalini and Liu, Sifei and Byeon, Wonmin and Breuel, Thomas and Kautz, Jan and Wang, Xiaolong},
  booktitle={CVPR},
  year={2022}
}

@inproceedings{ViT,
  title={An Image is Worth 16x16 Words: Transformers for Image Recognition at Scale},
  author={Dosovitskiy, Alexey and Beyer, Lucas and Kolesnikov, Alexander and Weissenborn, Dirk and Zhai, Xiaohua and Unterthiner, Thomas and Dehghani, Mostafa and Minderer, Matthias and Heigold, Georg and Gelly, Sylvain and others},
  booktitle={ICLR},
  year={2021}
}

@inproceedings{devlin2019bert,
  title={{BERT}: Pre-training of Deep Bidirectional Transformers for Language Understanding},
  author={Devlin, Jacob and Chang, Ming-Wei and Lee, Kenton and Toutanova, Kristina},
  booktitle={NAACL},
  year={2019}
}

@article{tarjan1975efficiency,
  title={Efficiency of a good but not linear set union algorithm},
  author={Tarjan, Robert Endre},
  journal={JACM},
  volume={22},
  number={2},
  pages={215--225},
  year={1975},
  publisher={ACM New York, NY, USA}
}

@inproceedings{flickr30kcn,
  title={Fluency-guided cross-lingual image captioning},
  author={Lan, Weiyu and Li, Xirong and Dong, Jianfeng},
  booktitle={ACM MM},
  pages={1549--1557},
  year={2017}
}

@article{cococn,
  title={COCO-CN for cross-lingual image tagging, captioning, and retrieval},
  author={Li, Xirong and Xu, Chaoxi and Wang, Xiaoxu and Lan, Weiyu and Jia, Zhengxiong and Yang, Gang and Xu, Jieping},
  journal={TMM},
  volume={21},
  number={9},
  pages={2347--2360},
  year={2019},
  publisher={IEEE}
}

@inproceedings{liu2024mmbench,
  title={Mmbench: Is your multi-modal model an all-around player?},
  author={Liu, Yuan and Duan, Haodong and Zhang, Yuanhan and Li, Bo and Zhang, Songyang and Zhao, Wangbo and Yuan, Yike and Wang, Jiaqi and He, Conghui and Liu, Ziwei and others},
  booktitle={ECCV},
  pages={216--233},
  year={2024},
  organization={Springer}
}

@article{zhang2024mme,
  title={Mme-realworld: Could your multimodal llm challenge high-resolution real-world scenarios that are difficult for humans?},
  author={Zhang, Yi-Fan and Zhang, Huanyu and Tian, Haochen and Fu, Chaoyou and Zhang, Shuangqing and Wu, Junfei and Li, Feng and Wang, Kun and Wen, Qingsong and Zhang, Zhang and others},
  journal={arXiv preprint arXiv:2408.13257},
  year={2024}
}

@article{zhang2024cmmmu,
  title={Cmmmu: A chinese massive multi-discipline multimodal understanding benchmark},
  author={Zhang, Ge and Du, Xinrun and Chen, Bei and Liang, Yiming and Luo, Tongxu and Zheng, Tianyu and Zhu, Kang and Cheng, Yuyang and Xu, Chunpu and Guo, Shuyue and others},
  journal={arXiv preprint arXiv:2401.11944},
  year={2024}
}

@article{fu2024ocrbench,
  title={Ocrbench v2: An improved benchmark for evaluating large multimodal models on visual text localization and reasoning},
  author={Fu, Ling and Kuang, Zhebin and Song, Jiajun and Huang, Mingxin and Yang, Biao and Li, Yuzhe and Zhu, Linghao and Luo, Qidi and Wang, Xinyu and Lu, Hao and others},
  journal={arXiv preprint arXiv:2501.00321},
  year={2024}
}

@misc{wang2022ofaunifyingarchitecturestasks,
      title={OFA: Unifying Architectures, Tasks, and Modalities Through a Simple Sequence-to-Sequence Learning Framework}, 
      author={Peng Wang and An Yang and Rui Men and Junyang Lin and Shuai Bai and Zhikang Li and Jianxin Ma and Chang Zhou and Jingren Zhou and Hongxia Yang},
      year={2022},
      eprint={2202.03052},
      archivePrefix={arXiv},
      primaryClass={cs.CV},
      url={https://arxiv.org/abs/2202.03052}, 
}

@misc{liu2024llavanext,
    title={LLaVA-NeXT: Improved reasoning, OCR, and world knowledge},
    url={https://llava-vl.github.io/blog/2024-01-30-llava-next/},
    author={Liu, Haotian and Li, Chunyuan and Li, Yuheng and Li, Bo and Zhang, Yuanhan and Shen, Sheng and Lee, Yong Jae},
    month={January},
    year={2024}
}

@misc{grootendorst2022bertopicneuraltopicmodeling,
      title={BERTopic: Neural topic modeling with a class-based TF-IDF procedure}, 
      author={Maarten Grootendorst},
      year={2022},
      eprint={2203.05794},
      archivePrefix={arXiv},
      primaryClass={cs.CL},
      url={https://arxiv.org/abs/2203.05794}, 
}

@misc{unmap_paper,
      title={UMAP: Uniform Manifold Approximation and Projection for Dimension Reduction}, 
      author={Leland McInnes and John Healy and James Melville},
      year={2020},
      eprint={1802.03426},
      archivePrefix={arXiv},
      primaryClass={stat.ML},
      url={https://arxiv.org/abs/1802.03426}, 
}

@inproceedings{cnbert,
  title={Revisiting Pre-Trained Models for Chinese Natural Language Processing},
  author={Cui, Yiming and Che, Wanxiang and Liu, Ting and Qin, Bing and Wang, Shijin and Hu, Guoping},
  booktitle={EMNLP},
  pages={657--668},
  year={2020}
}

@article{faiss,
  title={Billion-scale similarity search with {GPUs}},
  author={Johnson, Jeff and Douze, Matthijs and J{\'e}gou, Herv{\'e}},
  journal={IEEE Transactions on Big Data},
  volume={7},
  number={3},
  pages={535--547},
  year={2019},
  publisher={IEEE}
}

@inproceedings{kempf2025and,
  title={When and How Does CLIP Enable Domain and Compositional Generalization?},
  author={Kempf, Elias and Schrodi, Simon and Argus, Max and Brox, Thomas},
  booktitle = {ICML},
  year={2025}
}

@misc{docci_papper,
      title={DOCCI: Descriptions of Connected and Contrasting Images}, 
      author={Yasumasa Onoe and Sunayana Rane and Zachary Berger and Yonatan Bitton and Jaemin Cho and Roopal Garg and Alexander Ku and Zarana Parekh and Jordi Pont-Tuset and Garrett Tanzer and Su Wang and Jason Baldridge},
      year={2024},
      eprint={2404.19753},
      archivePrefix={arXiv},
      primaryClass={cs.CV},
      url={https://arxiv.org/abs/2404.19753}, 
}

@inproceedings{dci_paper,
  title={A picture is worth more than 77 text tokens: Evaluating clip-style models on dense captions},
  author={Urbanek, Jack and Bordes, Florian and Astolfi, Pietro and Williamson, Mary and Sharma, Vasu and Romero-Soriano, Adriana},
  booktitle={CVPR},
  year={2024}
}

@INPROCEEDINGS{caltech101,
  author={Li Fei-Fei and Fergus, R. and Perona, P.},
  booktitle={CVPRW}, 
  title={Learning Generative Visual Models from Few Training Examples: An Incremental Bayesian Approach Tested on 101 Object Categories}, 
  year={2004},
}

@article{cifar100,
  title={Learning multiple layers of features from tiny images},
  author={Krizhevsky, Alex and Hinton, Geoffrey and others},
  year={2009},
  publisher={Toronto, ON, Canada}
}

@misc{country211,
      title={Learning Transferable Visual Models From Natural Language Supervision}, 
      author={Alec Radford and Jong Wook Kim and Chris Hallacy and Aditya Ramesh and Gabriel Goh and Sandhini Agarwal and Girish Sastry and Amanda Askell and Pamela Mishkin and Jack Clark and Gretchen Krueger and Ilya Sutskever},
      year={2021},
      eprint={2103.00020},
      archivePrefix={arXiv},
      primaryClass={cs.CV},
      url={https://arxiv.org/abs/2103.00020}, 
}

@misc{DTD,
      title={Describing Textures in the Wild}, 
      author={Mircea Cimpoi and Subhransu Maji and Iasonas Kokkinos and Sammy Mohamed and Andrea Vedaldi},
      year={2013},
      eprint={1311.3618},
      archivePrefix={arXiv},
      primaryClass={cs.CV},
      url={https://arxiv.org/abs/1311.3618}, 
}

@inproceedings{Food101,
  title={Food-101--mining discriminative components with random forests},
  author={Bossard, Lukas and Guillaumin, Matthieu and Van Gool, Luc},
  booktitle={ECCV},
  pages={446--461},
  year={2014},
  organization={Springer}
}

@ARTICLE{MNIST,
  author={Lecun, Y. and Bottou, L. and Bengio, Y. and Haffner, P.},
  journal={Proceedings of the IEEE}, 
  title={Gradient-based learning applied to document recognition}, 
  year={1998},
}

@INPROCEEDINGS{Flowers102,
  author={Nilsback, Maria-Elena and Zisserman, Andrew},
  booktitle={2008 Sixth Indian Conference on Computer Vision, Graphics \& Image Processing}, 
  title={Automated Flower Classification over a Large Number of Classes}, 
  year={2008},
}

@INPROCEEDINGS{Pets,
  author={Parkhi, Omkar M and Vedaldi, Andrea and Zisserman, Andrew and Jawahar, C. V.},
  booktitle={CVPR}, 
  title={Cats and dogs}, 
  year={2012}
}

@ARTICLE{Resisc45,
  author={Cheng, Gong and Han, Junwei and Lu, Xiaoqiang},
  journal={Proceedings of the IEEE}, 
  title={Remote Sensing Image Scene Classification: Benchmark and State of the Art}, 
  year={2017},
}

@inproceedings{stanford_cars,
  title={3d object representations for fine-grained categorization},
  author={Krause, Jonathan and Stark, Michael and Deng, Jia and Fei-Fei, Li},
  booktitle={ICCVW},
  pages={554--561},
  year={2013}
}

@article{kiela2020hateful,
  title={The hateful memes challenge: Detecting hate speech in multimodal memes},
  author={Kiela, Douwe and Firooz, Hamed and Mohan, Aravind and Goswami, Vedanuj and Singh, Amanpreet and Ringshia, Pratik and Testuggine, Davide},
  journal={Neurips},
  year={2020}
}

@article{Voc2007,
  title={The pascal visual object classes (voc) challenge},
  author={Everingham, Mark and Van Gool, Luc and Williams, Christopher KI and Winn, John and Zisserman, Andrew},
  journal={IJCV},
  volume={88},
  number={2},
  pages={303--338},
  year={2010},
  publisher={Springer}
}

@inproceedings{wu2023grounded,
  title={Grounded image text matching with mismatched relation reasoning},
  author={Wu, Yu and Wei, Yana and Wang, Haozhe and Liu, Yongfei and Yang, Sibei and He, Xuming},
  booktitle={ICCV},
  pages={2976--2987},
  year={2023}
}
\bibliographystyle{colm2024_conference}
}
\newpage
\appendix

\section*{Appendix Overview}
The appendix includes the following sections:
\begin{itemize}[leftmargin=*, itemsep=0.2em, topsep=0.0em]
    \item \textbf{Appendix~\ref{appendix:Data construction}}: Provides detailed statistics of data filtration.
    \item \textbf{Appendix~\ref{appendix:examples}}: Visualizes image-text pair examples from the \dsname\ dataset.
    \item \textbf{Appendix~\ref{appendix:exploring}}: Offers additional analytical insights and exploratory studies, including source domain distributions (Sec.~\ref{appendix:top domains}), extended visualizations of topic modeling (Sec.~\ref{appendix: topic modeling}), and word cloud analyses (Sec.~\ref{appendix:word cloud}).
\end{itemize}

\section{Statistics of Data Filtration}
\label{appendix:Data construction}

\begin{table}[htbp]
\centering
\resizebox{\textwidth}{!}{
\begin{NiceTabular}{@{}llcccc@{}}
\toprule
\multicolumn{2}{c}{\textbf{Steps}} & \textbf{Left Data Num} & \textbf{Total Filter \%} & \textbf{Stage Filter \%} & \textbf{Left \%} \\ \midrule
\multicolumn{2}{l}{Collected image URL and text pairs} & 1,047,085,609 & -- & -- & 100.00\% \\ \midrule

\multirow{5}{*}{\shortstack[l]{Data \\ Source \\ Selection} } 
    & Language Control      & \multirow{3}{*}{726,334,674} & \multirow{3}{*}{30.63\%} & \multirow{3}{*}{69.37\%} & \multirow{3}{*}{69.37\%} \\
    & content Safety      &  &  &  &  \\
    & Source Reliability &  &  &  &  \\
    \cdashline{2-6}
    & Text Constraints & 706,069,936 & 32.57\% & 2.79\% & 67.43\% \\
    & Download Success & 475,104,485 & 54.63\% & 67.29\% & 45.37\% \\ \midrule

\multirow{8}{*}{\shortstack[l]{Text \\ Refinement}}
    & Linguistic Structure (CN-Detect) & 467,455,303 & 55.36\% & 1.61\% & 44.64\% \\
    & Linguistic Structure (Font conversion) & 467,455,303 & 55.36\% & 0.00\% & 44.64\% \\
    & Text Quality (Stop words) & 442,680,171 & 57.72\% & 5.30\% & 42.28\% \\
    & Text Quality (Nouns) & 431,878,774 & 58.75\% & 2.44\% & 41.25\% \\
    & Text Quality (\texttt{[UNK]}) & 431,619,647 & 58.78\% & 0.06\% & 41.22\% \\
    & Information Density (Entropy) & 400,068,251 & 61.79\% & 7.31\% & 38.21\% \\
    & Information Density (Emoji\&special chars) & 400,068,251 & 61.79\% & 0.00\% & 38.21\% \\
    & Content Safety & 397,187,759 & 62.07\% & 0.72\% & 37.93\% \\ \midrule

\multirow{6}{*}{\shortstack[l]{Visual \\ Diversification}}
    & Visual Fidelity & 352,663,012 & 66.32\% & 11.21\% & 33.68\% \\
    & Visual Fidelity (Low-textual images) & 352,204,550 & 66.36\% & 0.13\% & 33.64\% \\
    & Visual Fidelity (Blurry images) & 333,255,945 & 68.17\% & 5.38\% & 31.83\% \\
    & Information Density & 316,359,868 & 69.79\% & 5.07\% & 30.21\% \\
    & Perceptual and Semantic Redundancy & 186,019,602 & 82.23\% & 41.20\% & 17.77\% \\
    & Content Safety & 178,601,215 & 82.94\% & 3.25\% & 17.06\% \\ \midrule
    
\multirow{2}{*}{\shortstack[l]{Cross-Modal \\ Cross-Batch Filtering}} 
    & Cross-Modal Alignment Assessment & 154,293,590 & 85.26\% & 13.61\% & 14.74\% \\
    & Cross-Batch Redundancy Removal & 99,892,381 & 90.46\% & 35.26\% & 9.54\% \\ \midrule
\multicolumn{2}{l}{\textbf{Final image-text pairs}} & \textbf{99,892,381} & \textbf{90.46\%} & -- & \textbf{9.54\%} \\ \bottomrule
\end{NiceTabular}%
}
\vspace{-2mm}
\caption{Specific statistics information of the \dsname\ dataset construction pipeline.}
\vspace{-5mm}
\label{tab:data_filtering}
\end{table}

We further illustrate the statistical breakdown of our data construction pipeline in Tab.~\ref{tab:data_filtering}. Starting with an initial collection of approximately 1B raw image-text URLs, we apply a multi-stage filtering process, comprising data
source selection, text refinement, visual diversification, and cross-modal cross-batch filtering. This pipeline ultimately yields nearly 100M high-quality image-text pairs~(storage space occupies approximately 12TB), effectively filtering out 90\% of the original noise to ensure data quality.

\section{Examples in \dsname\ Dataset}
\label{appendix:examples}
\begin{figure*}[htbp]
    \centering
    \includegraphics[width=1\linewidth]{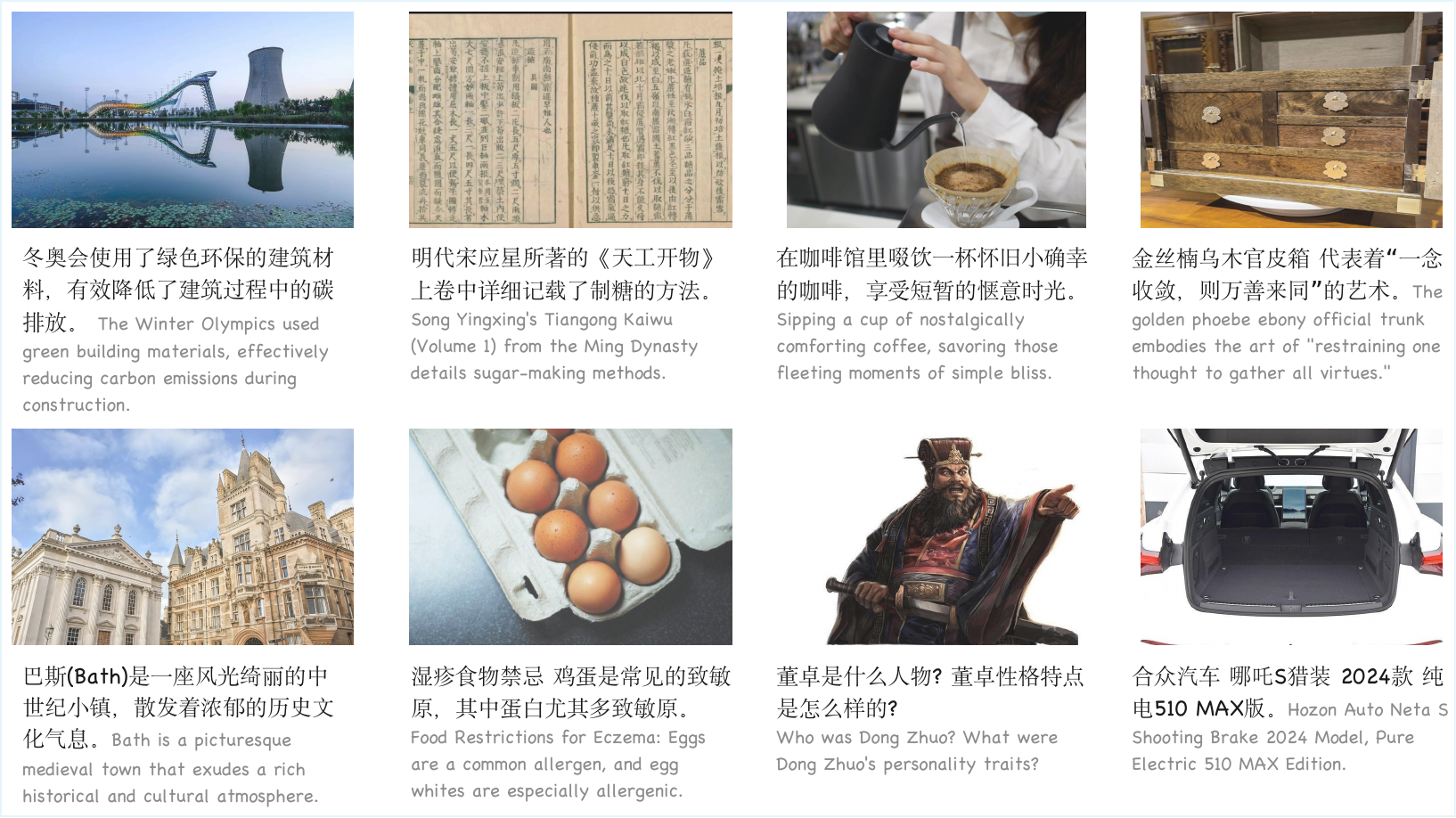}
    \vspace{-5mm}
    \caption{Visualization of image-text pairs in \dsname\ dataset.}
    \vspace{-3mm}
    \label{appendix_fig: data visualization}
\end{figure*}

Fig.~\ref{appendix_fig: data visualization} presents representative examples from the \dsname\ dataset, comprising images and their corresponding Chinese textual descriptions. Specifically, these image-text pairs encompass a wide range of domains, such as natural scenery, historical literature, and automotive technology, demonstrating the thematic diversity of our dataset. This breadth makes \dsname\ particularly well-suited for general-purpose Vision-Language Pre-training.

\section{Exploring \dsname}
\label{appendix:exploring}
\subsection{Source Domain Distribution}
\label{appendix:top domains}
\begin{table}[htbp]
\centering

\begin{NiceTabular}{clrrl}
\toprule
\textbf{Rank} & \textbf{Source Domain} & \textbf{Total Count} & \textbf{\%} & \textbf{Categories} \\
\midrule

1 & alicdn.com & 3,555,967 & 3.56\% & E-commerce, Cloud Computing \\
2 & baidu.com & 2,649,928 & 2.65\% & Search Engine, Tech \\
3 & wp.com & 1,582,010 & 1.58\% & Blog Hosting (WordPress), CMS \\
4 & aliyuncs.com & 961,433 & 0.96\% & Cloud Computing, Object Storage \\
5 & chem17.com & 781,994 & 0.78\% & Chemical Instruments, E-commerce \\
6 & bing.net & 741,185 & 0.74\% & Search Engine~(Bing), Content Delivery \\
7 & udn.com.tw & 542,113 & 0.54\% & News Media, General Info \\
8 & faiusr.com & 490,843 & 0.49\% & Website Builder, Marketing \\
9 & sinaimg.cn & 404,076 & 0.40\% & Social Media~(Weibo), Sina Portal \\
10 & wixstatic.com & 393,168 & 0.39\% & Website Builder~(Wix), Image Hosting \\
11 & 126.net & 372,122 & 0.37\% & Internet Services~(NetEase), Email \\
12 & 360buyimg.com & 336,282 & 0.34\% & E-commerce~(JD.com), Logistics \\
13 & vjshi.com & 333,778 & 0.33\% & Video Assets, Copyright Trading \\
14 & bing.com & 322,193 & 0.32\% & Search Engine, Microsoft, International \\
15 & staticflickr.com & 319,134 & 0.32\% & Photo Community, Social Media \\
16 & xiniu.com & 286,491 & 0.29\% & Enterprise Services, Marketing \\
17 & sohu.com & 285,513 & 0.29\% & General Portal, Media, Video \\
18 & myqcloud.com & 262,140 & 0.26\% & Cloud Computing~(Tencent) \\
19 & smzdm.com & 258,988 & 0.26\% & Consumer Guide, E-commerce \\
20 & made-in-china.com & 256,439 & 0.26\% & Cross-border Trade, B2B, Export \\
21 & cloudfront.net & 249,402 & 0.25\% & Content Delivery, Cloud Computing \\
22 & sogoucdn.com & 231,903 & 0.23\% & Search Engine~(Sogou) \\
23 & toutiaoimg.com & 229,255 & 0.23\% & News~(Toutiao), ByteDance \\
24 & hbzhan.com & 227,981 & 0.23\% & Environmental Industry \\
25 & byteimg.com & 225,890 & 0.23\% & ByteDance, Short Video, News \\
26 & qpic.cn & 223,192 & 0.22\% & Social Media~(QQ/WeChat), Tencent \\
27 & myapp.com & 216,009 & 0.22\% & App Store, Mobile Internet \\
28 & zdmimg.com & 210,646 & 0.21\% & Consumer Community \\
29 & tom.com & 210,437 & 0.21\% & General Portal, Internet Services, Media \\
30 & itc.cn & 204,060 & 0.20\% & Sohu Media, Content Delivery~(CDN) \\
31 & 588ku.com & 194,423 & 0.19\% & Design Materials~(Qinku/Wotu) \\
32 & qunarzz.com & 193,114 & 0.19\% & Travel \& Tourism~(Qunar) \\
33 & bdxiguaimg.com & 185,956 & 0.19\% & Video~(Xigua Video), ByteDance \\
34 & suning.cn & 174,229 & 0.17\% & E-commerce~(Suning), Retail \\
35 & thefastimg.com & 168,851 & 0.17\% & Image Hosting, General CDN \\
36 & gtimg.com & 168,577 & 0.17\% & Tencent, Games, Social \\
37 & iqiyipic.com & 167,389 & 0.17\% & Video~(iQIYI), Entertainment \\
38 & myxypt.com & 164,791 & 0.17\% & Pharmaceutical B2B, Industry Platform \\
39 & book.com.tw & 161,625 & 0.16\% & Book Retail, E-commerce, Culture \\
40 & tripcdn.com & 155,944 & 0.16\% & Travel~(Ctrip/Trip.com), International \\
\bottomrule
\end{NiceTabular}
\caption{Statistics and categories overview of the top 40 image source domains.}
\label{tab:domain_stats}
\end{table}

To further investigate the origins of our data, we identified and ranked the top 40 primary web sources, as detailed in Tab.~\ref{tab:domain_stats}. The results indicate that the majority of image-text pairs originate from widely-used Chinese platforms and applications, such as Alibaba, Baidu, ByteDance, and so on. Furthermore, these sources span diverse categories, including E-commerce, news media, and search engines, demonstrating the heterogeneous nature of our data. This variety validates that \dsname\ is sourced from a broad spectrum of real-world scenarios, capturing the richness of daily-life multimodal content.

\subsection{Topic Modeling}
\label{appendix: topic modeling}
\begin{figure*}[htbp]
    \centering
    \includegraphics[width=1\linewidth]{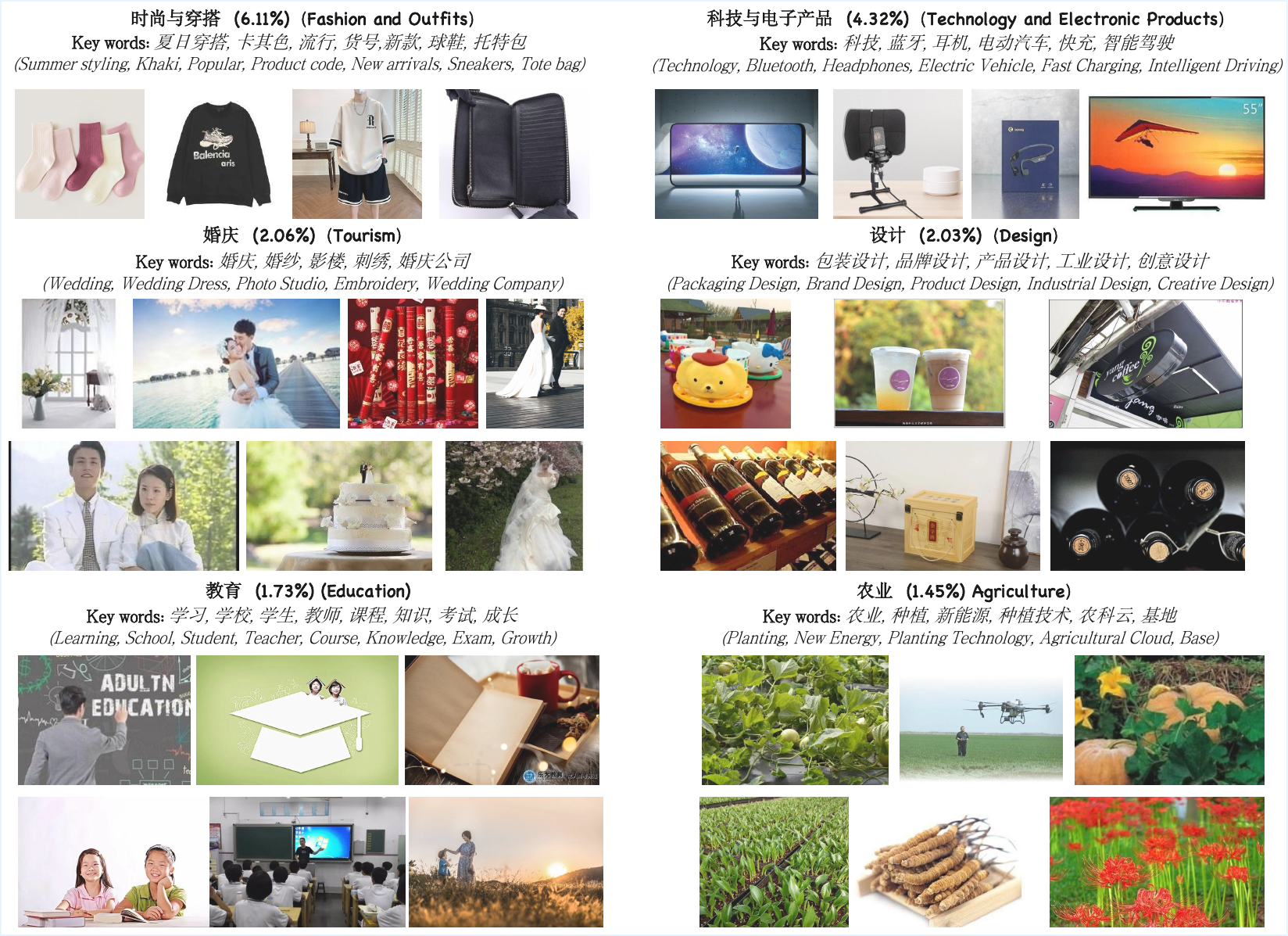}
    \vspace{-7mm}
    \caption{External topic examples visualization of \dsname\ dataset generated via the BERTopic~\cite{grootendorst2022bertopicneuraltopicmodeling}.}
    \label{fig:appendix topic}
\end{figure*}

We further illustrate the topic distribution of \dsname\ in Fig.~\ref{fig:appendix topic}, following the same analytical methodology described in Sec.~\ref{subsec: statistic of danqing}. The visualized examples encompass both previously mentioned categories and novel domains, such as tourism, design, education, and agriculture, which are prevalent throughout the dataset. This thematic breadth, exemplified by diverse visual content, underscores how \dsname\ aligns closely with real-world scenarios and everyday life.

\subsection{Word Cloud}
\label{appendix:word cloud}
\begin{figure*}[htbp]
    \centering
    \includegraphics[width=1\linewidth]{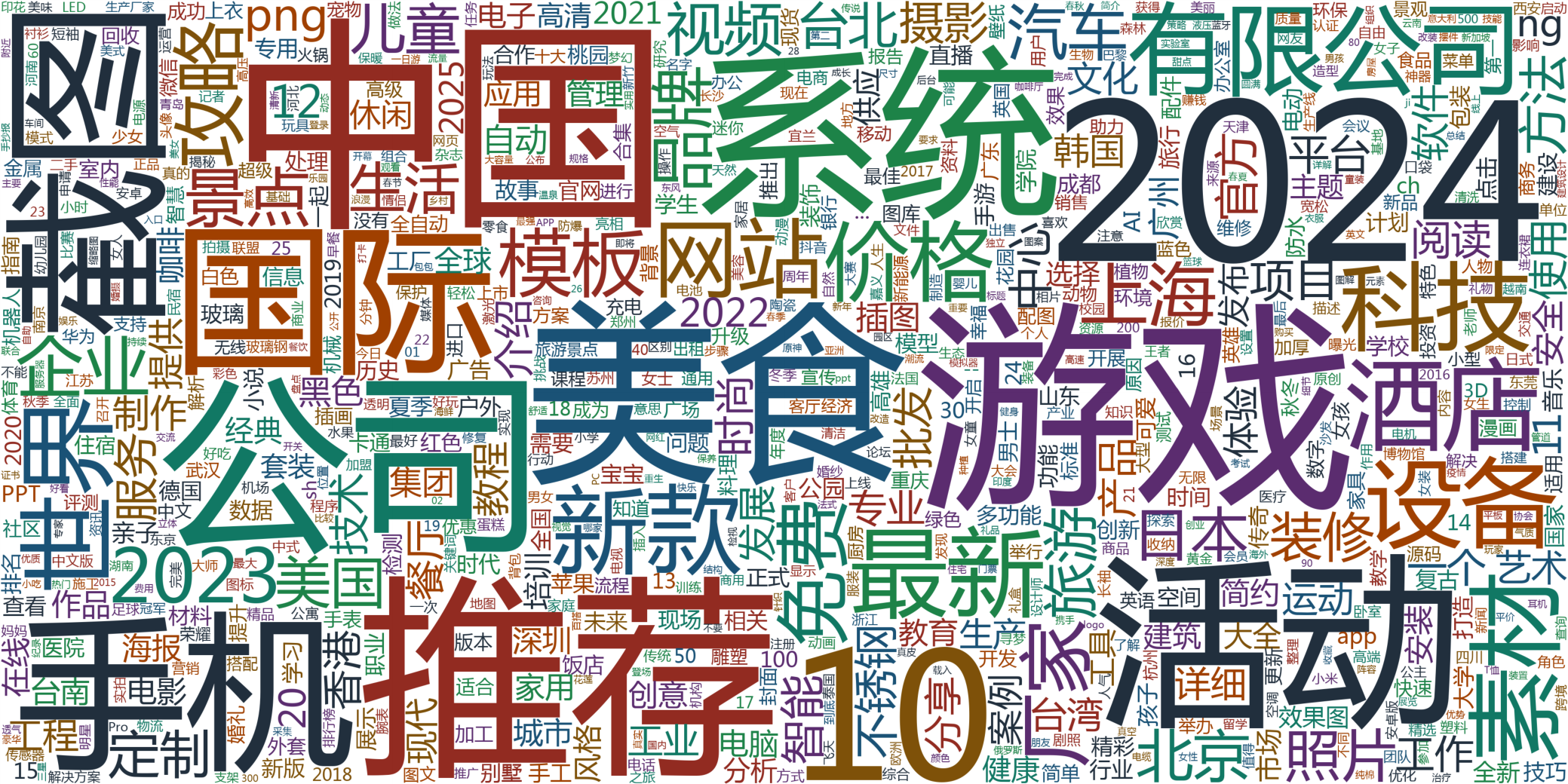}
    \vspace{-7mm}
    \caption{Word cloud visualization of 10M subset texts from \dsname.}
    \label{appendix_fig:word cloud}
\end{figure*}

In Fig.~\ref{appendix_fig:word cloud}, we visualize the distribution of Chinese words (comprising one or more tokens) within the \dsname\ dataset. Specifically, we utilize the jieba~\footnote{https://github.com/fxsjy/jieba} words segmentation module to tokenize the text and generate the word cloud. The visualization highlights that the most frequent terms include ''2024'', \begin{CJK*}{UTF8}{gkai}''中国''\end{CJK*} (China), \begin{CJK*}{UTF8}{gkai}''游戏''\end{CJK*} (Game), \begin{CJK*}{UTF8}{gkai}``美食''\end{CJK*} (Food), \begin{CJK*}{UTF8}{gkai}``活动''\end{CJK*} (Activity), and so on. This distribution demonstrates the inclusion of the newest semantic concepts and diverse daily topics, which are essential for robust, general purpose Vision-Language Pre-training.

\end{document}